%% file: paper.tex
\documentclass[10pt,twocolumn,letterpaper]{article}

\usepackage{cvpr}
\usepackage{times}
\usepackage{epsfig}
\usepackage{graphicx}
\usepackage{amsmath}
\usepackage{amssymb}

\usepackage{booktabs}
\usepackage{footnote}
\usepackage{graphicx}
\usepackage{setspace}
\usepackage{verbatim}

\usepackage[pagebackref=true,breaklinks=true,letterpaper=true,colorlinks,bookmarks=false]{hyperref}

\cvprfinalcopy %

\ifcvprfinal\fi
\begin{document}

\newcommand*{\comb}[2]{{}^{#1}C_{#2}}%
\newcommand*\rot{\rotatebox{90}}%

\title{Wider or Deeper: Revisiting the ResNet Model for Visual Recognition\thanks{Correspondence
should be addressed to C. Shen.}}

\author{Zifeng Wu, Chunhua Shen, and Anton van den Hengel\\
    School of Computer Science, The University of Adelaide, 
Adelaide, SA 5005, Australia\\
e-mail: {\tt  firstname.lastname@adelaide.edu.au}
}

\maketitle
\begin{abstract}
The trend towards increasingly deep neural networks has been driven by a general observation that increasing depth increases the performance of a network.
Recently, however, evidence has been amassing that simply increasing depth may not be the best way to increase performance, particularly given other limitations.
Investigations into deep residual networks have also suggested that they may not in fact be operating as a single deep network, but rather as an ensemble of many relatively shallow networks.
We examine these issues, and in doing so arrive at a new interpretation of the unravelled view of deep residual networks which explains some of the behaviours that have been observed experimentally.
As a result, we are able to derive a new, shallower, architecture of residual networks which significantly outperforms much deeper models such as \mbox{ResNet-200} on the ImageNet classification dataset.
We also show that this performance is transferable to other problem domains by developing a semantic segmentation approach which outperforms the state-of-the-art by a remarkable margin on datasets including PASCAL VOC, PASCAL Context, and Cityscapes.
The architecture that we propose thus outperforms its comparators, including very deep ResNets, and yet is more efficient in memory use and sometimes also in training time.
The code and models are available at \url{https://github.com/itijyou/ademxapp}.

\end{abstract}

\tableofcontents
\clearpage

\section{Introduction}

The convolutional networks used by the computer vision community have been growing deeper and deeper each year since Krizhevsky~et~al.~\cite{AlexNet.NIPS.2012.Krizhevsky} proposed AlexNet in 2012.
The deepest network~\cite{ResNet.CVPR.2016.He} in the literature is a residual network~(ResNet) with 1,202 trainable layers, which was trained using the tiny images in the CIFAR-10 dataset~\cite{CIFAR10.2009.Krizhevsky}.
The image size here is important, because it means that the size of corresponding feature maps is relatively small, which is critical in training extremely deep models.
Most networks operating on more practically interesting image sizes tend to have the order of one, to two, hundred layers, \eg
the 200-layer ResNet~\cite{ResNet2.2016.He} and 96-layer Inception-ResNet~\cite{InceptionResNet.2016.Szegedy}.
The progression to deeper networks continues, however, with Zhao~et~al.~\cite{SenseCUSceneParsing.2016.Zhao} having trained a 269-layer network for semantic image segmentation.
These networks were trained using the ImageNet classification dataset~\cite{ILSVRC2015.Russakovsky}, where the images are of much higher resolution.
Each additional layer requires not only additional memory, but also additional training.
The marginal gains achieved by each additional layer diminish with depth, however, to the point where
Zhao~et~al.~\cite{SenseCUSceneParsing.2016.Zhao} achieved only an improvement of 1.1\% (from 42.2\% to 43.3\% by mean intersection-over-union scores)
after almost doubling the number of layers~(from 152 to 269).
On the other hand, Zagoruyko and Komodakis showed that it is possible to train much shallower but wider networks on CIFAR-10, which outperform a ResNet\cite{ResNet.CVPR.2016.He} with its more than one thousand layers.
The question thus naturally arises as to whether deep, or wide, is the right strategy.

\begin{figure*}[t]
\begin{center}
\includegraphics[width=0.6798\linewidth,trim=20 260 150 0]{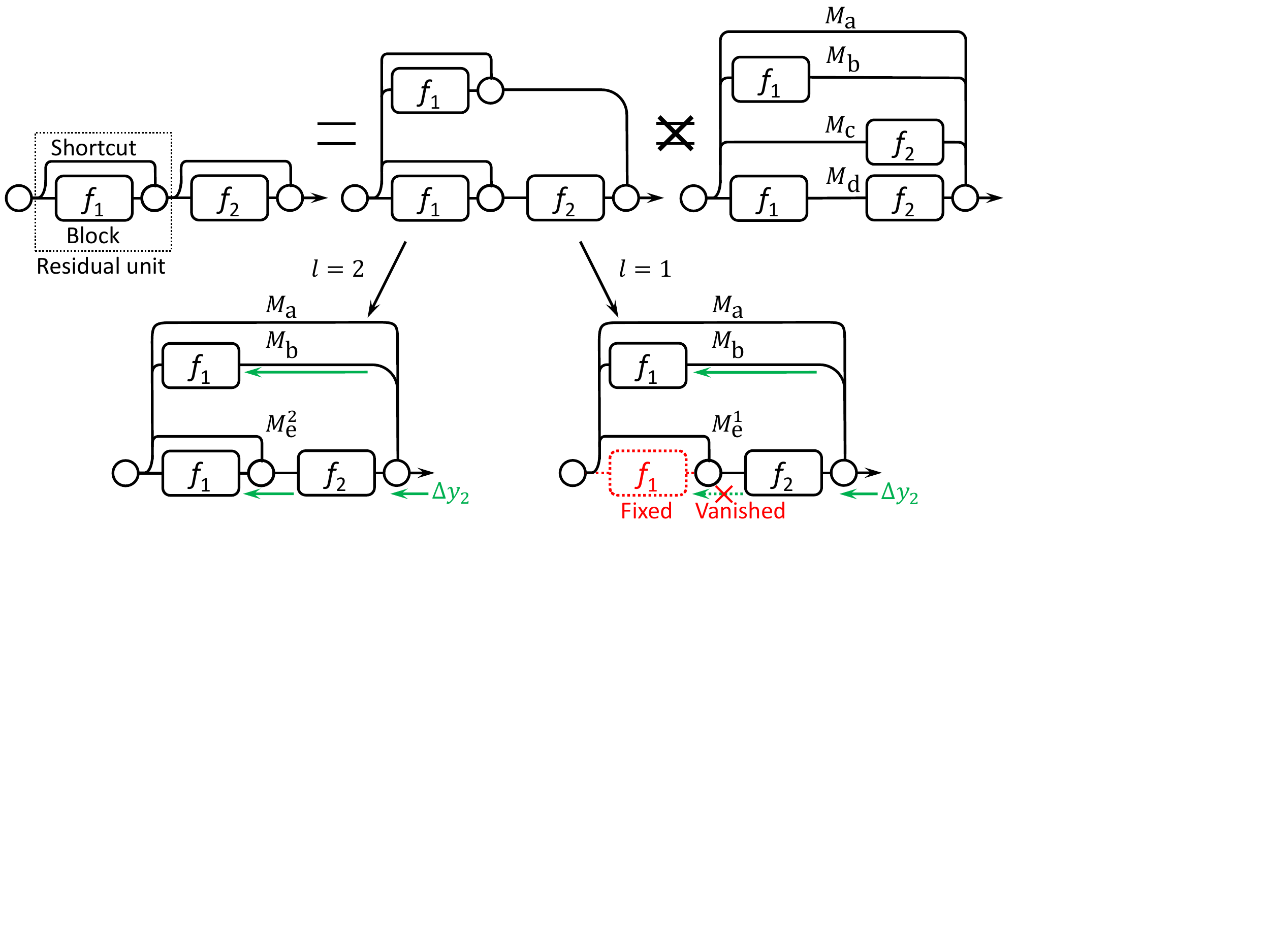}
\end{center}
\caption{
The unravelled view of a simple ResNet.
The fact that $f_2(\cdot)$ is non-linear gives rise to the inequality in the top row, as $f_2(a+b)\neq f_2(a)+ f_2(b)$.
This shows that $f_2(\cdot)$ never operates independently of the result of $f_1(\cdot)$, and thus that the number of independent classifiers increases linearly with the number of residual units.
Analysing the interactions between residual units at a given effective depth, here labelled $l$, illuminates the paths taken by gradients during training.
}
\label{fig:unraveled}
\end{figure*}

In order to examine the issue we first need to understand the mechanism behind ResNets.
Veit~et~al.~\cite{UnraveledResNet.2016.Veit} have claimed that they actually behave as exponential ensembles of relatively shallow networks.
However, there is a gap between their proposed unravelled view of a ResNet, and a real exponential ensemble of sub-networks, as illustrated in the top row of Fig.~\ref{fig:unraveled}.
Since the residual units are non-linear,
we cannot further split the bottom path into two sub-networks, i.e., $M_\textrm{c}$ and $M_\textrm{d}$.
It turns out that ResNets are only assembling linearly growing numbers of sub-networks.
Besides, the key characteristic of our introduced view is that it depends on the effective depth $l$ of a network.
This $l$ amounts to the number of residual units which backward gradients during training can go through.
When $l \ge 2$,
the two-unit ResNet in Fig.~\ref{fig:unraveled} can be seen as an ensemble of three sub-networks,
i.e., $M_\textrm{a}$, $M_\textrm{b}$, and $M_\textrm{e}^2$,
as shown in the bottom left.
When $l = 1$,
nothing changes except that we replace the third sub-network with a shallower one $M_\textrm{e}^1$,
as shown in the bottom right example.
The superscripts in $M_\textrm{e}^1$ and $M_\textrm{e}^2$ denote their actual depths.
About the unravelled view, the effective depth of a ResNet, and the actual depth of a sub-network, more details will be provided in the sequence.
It is also worth noting that Veit~et~al.~\cite{UnraveledResNet.2016.Veit} empirically found that most gradients in a 110-layer ResNet can only go through up to seventeen residual units,
which supports our above hypothesis that the effective depth~$l$ exists for a specific network.

In this paper, our contributions include:
\begin{itemize}
\item
We introduce a further developed intuitive view of ResNets,
which helps us understand their behaviours,
and find possible directions to further improvements.
\item
We propose a group of relatively shallow convolutional networks based on our new understanding.
Some of them achieve the state-of-the-art results on the ImageNet classification dataset~\cite{ILSVRC2015.Russakovsky}.
\item
We evaluate the impact of using different networks on the performance of semantic image segmentation,
and show these networks, as pre-trained features, can boost existing algorithms a lot.
We achieve the best results on PASCAL VOC~\cite{PascalVoc.IJCV.2014.Everingham}, PASCAL Context~\cite{PascalContext.CVPR.2014.Mottaghi}, and Cityscapes~\cite{Cityscapes.CVPR.2016.Cordts}.
\end{itemize}

\section{Related work}

Our work here is closely related to two topics, residual network~(ResNet) based image classification and semantic image segmentation using fully convolutional networks .

As we have noted above, He~et~al.~\cite{ResNet.CVPR.2016.He,ResNet2.2016.He} recently proposed the ResNets to combat the vanishing gradient problem during training very deep convolutional networks.
ResNets have outperformed previous models at a variety of tasks, such as object detection~\cite{MNC.CVPR.2016.Dai} and semantic image segmentation~\cite{DeepLab2.2016.Chen}.
They are gradually replacing VGGNets~\cite{VGGNet.2014.Simonyan} in the computer vision community, as the standard feature extractors.
Nevertheless, the real mechanism underpinning the effectiveness of ResNets is not yet clear.
Veit~et~al.~\cite{UnraveledResNet.2016.Veit} claimed that they behave like exponential ensembles of relatively shallow networks,
yet the `exponential' nature of the ensembles has yet to be theoretically verified.
Residual units are usually non-linear, which prevents a ResNet from exponentially expanding into separated sub-networks,
as illustrated in Fig.~\ref{fig:unraveled}.
It is also unclear as to whether a residual structure is required to train very deep networks.
For example, Szegedy~et~al.~\cite{InceptionResNet.2016.Szegedy} showed that it is `not very difficult' to train competitively deep networks, even without residual shortcuts.
Currently, the most clear advantage of ResNets is in their fast convergence~\cite{ResNet.CVPR.2016.He}.
Szegedy~et~al.~\cite{InceptionResNet.2016.Szegedy} observed similar empirically results to support that.
On the other hand, Zagoruyko and Komodakis~\cite{WRN.2016.Zagoruyko} found that a wide sixteen-layer ResNet outperformed the original thin thousand-layer ResNet~\cite{ResNet2.2016.He} on datasets composed of tiny images such as CIFAR-10~\cite{CIFAR10.2009.Krizhevsky}.
The analysis we present here is motivated by their empirical testing, but aims at a more theoretical approach, and the observation that a grid search of configuration space is impractical on large scale datasets such as the ImageNet classification dataset~\cite{ILSVRC2015.Russakovsky}.

Semantic image segmentation amounts to predicting the categories for each pixel in an image.
Long~et~al.~\cite{FCN.CVPR.2015.Long} proposed the fully convolutional networks (FCN) to this end.
FCNs soon became the mainstream approach to dense prediction based tasks, especially due to its efficiency.
Besides, empirical results in the literature~\cite{DeepLab2.2016.Chen} showed that stronger pre-trained features can yet further improve their performance.
We thus here base our semantic image segmentation approach on fully convolutional networks, and will show the impact of different pre-trained features on final segmentation results.

\section{Residual networks revisited}

We are concerned here with
the full pre-activation version of residual networks (ResNet)~\cite{ResNet2.2016.He}.
For shortcut connections, we consider identity mappings~\cite{ResNet2.2016.He} only.
We omit the raw input and the top-most linear classifier for clarity.
Usually, there may be a stem block~\cite{InceptionResNet.2016.Szegedy} or 
several traditional
convolution layers~\cite{ResNet.CVPR.2016.He,ResNet2.2016.He} directly after the raw input.
We omit these also, for the purpose of simplicity.

For the residual Unit $i$, let $y_{i-1}$ be the input,
and let $f_i(\cdot)$ be its trainable non-linear mappings, also named Block~$i$.
The output of Unit $i$ is recursively defined as:
\begin{equation}\label{eqn:residual}
y_i \equiv f_i(y_{i-1}, \boldsymbol{w}_i) + y_{i-1},
\end{equation}
where $\boldsymbol{w}_i$ denotes the trainable parameters,
and $f_i(\cdot)$ is often two or three stacked convolution stages.
In the full pre-activation version, the components of a stage are in turn a batch normalization~\cite{BatchNorm.2015.Ioffe}, a rectified linear unit~\cite{ReLU.ICML.2010.Nair}~(ReLU) non-linearity, and a convolution layer.

\subsection{Residual networks unravelled online}

Applying Eqn.(\ref{eqn:residual}) in one substitution step, we expand the forward pass into:
\begin{eqnarray}
\!\!\!\!\!\! y_2 & \!\!\!\!=\!\!\!\! & y_1 \!+\! f_2(y_1,\! \boldsymbol{w}_2) \\
\!\!\!\!\!\! & \!\!\!\!=\!\!\!\! & y_0 \!+\! f_1(y_0,\! \boldsymbol{w}_1) \!+\! f_2(y_0 \!+\! f_1(y_0,\! \boldsymbol{w}_1),\! \boldsymbol{w}_2) \label{eqn:unraveled} \\
\!\!\!\!\!\! & \!\!\!\!\neq\!\!\!\! & y_0 \!+\! f_1(y_0,\! \boldsymbol{w}_1) \!+\! f_2(y_0,\! \boldsymbol{w}_2) \!+\! f_2(f_1(y_0,\! \boldsymbol{w}_1),\! \boldsymbol{w}_2), \label{eqn:full ensemble}
\end{eqnarray}
which describes the unravelled view by Veit~et~al.~\cite{UnraveledResNet.2016.Veit}, as shown in the top row of Fig.~\ref{fig:unraveled}.
Since $f_2(\cdot)$ is non-linear, we cannot derive Eqn.(\ref{eqn:full ensemble}) from Eqn.(\ref{eqn:unraveled}).
So the whole network is not equivalent to an exponentially growing ensemble of sub-networks.
It is rather, more accurately, described as a linearly growing ensemble of sub-networks.
For the two-unit ResNet as illustrated in~Fig.~\ref{fig:unraveled}, there are three, e.g., $M_\textrm{a}$, $M_\textrm{b}$, and $M_\textrm{e}^2$,
sub-networks respectively corresponding to the three terms in Eqn.(\ref{eqn:unraveled}), i.e., $y_0$, $f_1(y_0, \boldsymbol{w}_1)$, and \mbox{$f_2(y_0 + f_1(y_0, \boldsymbol{w}_1), \boldsymbol{w}_2)$}.

Veit~et~al.~in~\cite{UnraveledResNet.2016.Veit} showed that the paths which gradients take through a ResNet are typically far shorter than the total depth of that network.
They thus introduced the idea of \emph{effective depth} as a measure for the true length of these paths.  By characterising the units of a ResNet given its effective depth,
we illuminate the impact of varying paths that gradients actually take,
as in Fig.~\ref{fig:unraveled}.
We here illustrate this impact in terms of small effective depths,
because to do so for larger ones would require diagrams of enormous networks.
The impact is the same, however.

Take the ResNet in~Fig.~\ref{fig:unraveled} for example again. 
In an SGD iteration,
the backward gradients are:
\begin{eqnarray}
\Delta \boldsymbol{w}_2 & = & \frac{\textrm{d} f_2}{\textrm{d} \boldsymbol{w}_2} \cdot \Delta y_2 \\
\Delta y_1 & = & \Delta y_2 + f_2' \cdot \Delta y_2 \\
\Delta \boldsymbol{w}_1 & = & \frac{\textrm{d} f_1}{\textrm{d} \boldsymbol{w}_1} \cdot \Delta y_2 + \frac{\textrm{d} f_1}{\textrm{d} \boldsymbol{w}_1} \cdot f_2' \cdot \Delta y_2, \label{eqn:unravel online dw1}
\end{eqnarray}
where $f_2'$ denotes the derivative of $f_2(\cdot)$ to its input $y_1$.
When effective depth $l \ge 2$, both terms in~Eqn.(\ref{eqn:unravel online dw1}) are non-zeros,
which corresponds to the bottom-left case in~Fig.~\ref{fig:unraveled}.
Namely, Block~1 receives gradients from both $M_\textrm{b}$ and $M_\textrm{e}^2$.
However, when effective depth $l = 1$, the gradient $\Delta y_2$ vanishes after passing through Block~2.
Namely, $f_2' \cdot \Delta y_2 \rightarrow 0$.
So, the second term in~Eqn.(\ref{eqn:unravel online dw1}) also goes to zeros,
which is illustrated by the bottom-right case in~Fig.~\ref{fig:unraveled}.
The weights in Block~1 indeed vary across different iterations,
but they are updated only by $M_\textrm{b}$.
To $M_\textrm{e}^1$, Block~1 is no more than an additional input providing preprocessed representations,
because Block~1 is not end-to-end trained, from the point of view of $M_\textrm{e}^1$.
In this case, we name $M_\textrm{e}^1$ to have an \emph{actual depth} of one.
We say that the ResNet is over-deepened,
and that it cannot be trained in a fully end-to-end manner, even with those shortcut connections.

\begin{figure}[t]
\begin{center}
\includegraphics[width=0.90\linewidth,trim=0 190 180 10]{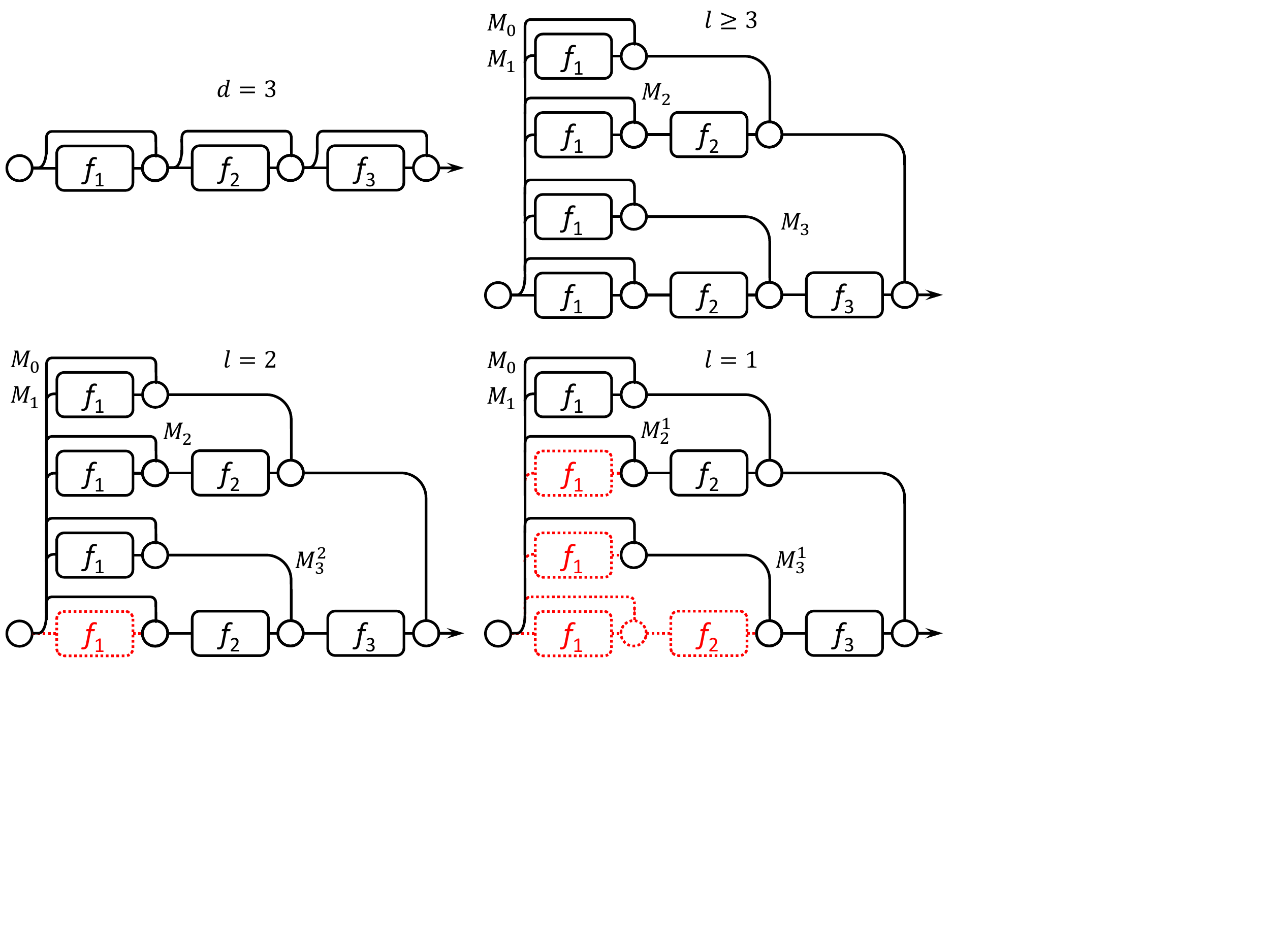}
\end{center}
\caption{The impact of inserting an extra residual unit into a two-unit ResNet,
which depends on the effective depth $l$.}
\label{fig:unraveled_online}
\end{figure}

Let $d$ be the total number of residual units.
We can see a ResNet as an ensemble of different sub-networks, i.e., $M_i, i = \{0, 1, \cdots, d\}$.
The actual depth of $M_i$ is $\min (i, l)$.
We show an unravelled three-unit ResNet with different effective depths in~Fig.~\ref{fig:unraveled_online}.
By way of example, note that $M_1$ in Fig.~\ref{fig:unraveled_online} contains only Block~1,
whereas $M_2$ contains both Block~1~and~Block~2.
Among the cases illustrated, the bottom left example is more complicated, where $d=3$ and $l=2$.
From the point of view of $M_3^2$, the gradient of Block~1 is
$f_3' \cdot \Delta y_3 + f_2' \cdot f_3' \cdot \Delta y_3$,
where the first term is non-zero.
$M_3^2$ will thus update Block~1 at each iteration.
Considering the non-linearity in Block~3, it is non-trivial to tell if this is as good as the fully end-to-end training case, as illustrated by $M_3$ in the top right example.
An investigation of this issue remains future work.

\subsection{Residual networks behaviours revisited}

\textbf{Very deep ResNets}.
Conventionally, it is not easy to train very deep networks due to the vanishing gradient problem~\cite{Deep.TNN.1994.Bengio}.
To understand how a very deep ResNet is trained,
the observation by~Veit~et~al.~\cite{UnraveledResNet.2016.Veit} is important,
i.e., gradients vanish exponentially as the length of paths increases.
Now refer to the top-right example in~Fig.~\ref{fig:unraveled_online}.
This is somewhat similar to the case of a shallow or reasonably deep ResNet, when $d \le l$.
At the beginning, the shallowest sub-network,~i.e., $M_1$, converges fast,
because it gives Block~1 the largest gradients.
From the point of view of $M_2$, Block~2 may also receive large gradients due to the path with a length of one.
However, the input of Block~2 partly depends on Block~1.
It would not be easy for Block~2 to converge before the output of Block~1 stabilises.
Similarly, Block~3 will need to wait for Blocks~1~and~2, and so forth.
In this way, a ResNet seems like an ensemble with a growing number of sub-networks.
Besides, each newly added sub-network will have a larger actual depth than all the previous ones.
Note that Littwin and Wolf~\cite{ResNetsAreEnsembles.2016.Littwin}, in a concurrent work,
have theoretically showed that ResNets are virtual ensembles whose depth grows as training progresses.
Their result to some extent coincides with the above described process.

The story will however be different when the actual depth becomes as large as the effective depth.
Refer to the bottom-right example in~Fig.~\ref{fig:unraveled_online}.
This is somewhat similar to the case of an over-deepened ResNet, when $d$ is much larger than $l$.
Again, Block~1 in $M_1$ gets trained and stabilises first.
However, this time $M_2^1$ is not fully end-to-end trained any more.
Since $M_2^1$ gives no gradients to Block~1,
it becomes a one-block sub-network trained on top of some preprocessed representations,
which are obtained by adding the output of Block~1 up to the original input.
In this way, the newly added sub-network $M_2^1$ still has an actual depth of one,
which is no deeper than the previous one, i.e., $M_1$, and so forth for $M_3^1$.
ResNets thus avoid the vanishing gradient problem by reshaping themselves into multiple shallower sub-networks.
This is just another view of delivering gradients to bottom layers through shortcut connections.
Researchers~\cite{FractalNet.2016.Larsson,InceptionResNet.2016.Szegedy} have claimed that the residual shortcut connections are not necessary even in very deep networks.
However, there are usually short paths in their proposed networks as well.
For example, the 76-layer Inception-v4 network~\cite{InceptionResNet.2016.Szegedy} has a much shorter twenty-layer route from its input to the output.
There might be differences in the details~\cite{DeepFuse.2016.Wang} between fusion by concatenation~(Inception-v4) and fusion by summation~(ResNets).
However, the manner of avoiding the vanishing gradient problem is probably similar,~i.e., using shortcuts, either with trainable weights or not.
We are thus not yet in a position to be able to claim that the vanishing gradient problem has been solved.

\textbf{Wide ResNets}.
Conventionally, wide layers are more prone to over-fitting,
and sometimes require extra regularization such as dropout~\cite{Dropout.JMLR.2014.Srivastava}.
However, Zagoruyko and Komodakis~\cite{WRN.2016.Zagoruyko} showed the possibility to effectively train times wider ResNets, even without any extra regularization.
To understand how a wide ResNet is trained,
refer to the top right example in~Fig.~\ref{fig:unraveled_online}.
This simulates the case of a rather shallow network, when $d$ is smaller than $l$.
We reuse the weights of Block~1 for four times.
Among these, Block~1 is located in three different kinds of circumstances.
In the bottom-most path of the sub-network~$M_3$, it is supposed to learn some low-level features;
in $M_2$, it should learn both low-level and mid-level features;
and in $M_1$, it has to learn everything.
This format of weight sharing may suppress over-fitting,
especially for those units far from the top-most linear classifier.
Hence ResNets inherently introduce regularization by weight sharing among multiple very different sub-networks.

\textbf{Residual unit choices}.
For better performance,
we hope that a ResNet should expand into a sufficiently large number of sub-networks,
some of which should have large model capacity.
So, given our previous observations, the requirements for an ideal mapping function in a residual unit are,
1) being strong enough to converge even if it is reused in many sub-networks, and
2) being shallow enough to enable an large effective depth.
Since it is very hard to build a model with large capacity using a single trainable layer~\cite{ResNet2.2016.He}, the most natural choice would be a residual unit with two wide convolution stages.
This coincides with empirical results reported by Zagoruyko and Komodakis~\cite{WRN.2016.Zagoruyko}.
They found that, among the most trivial structure choices, the best one is to stack two $3 \times 3$ convolution stages.

\subsection{Wider or deeper?}
\vspace{-2.0mm}

To summarize the previous subsections, shortcut connections enable us to train wider and deeper networks.
As they growing to some point, we will face the dilemma between width and depth.
From that point, going deep, we will actually get a wider network,
with extra features which are not completely end-to-end trained;
going wider, we will literally get a wider network,
without changing its end-to-end characteristic.
We have learned the strength of depth from the previous plain deep networks without any shortcuts, e.g., the AlexNet~\cite{AlexNet.NIPS.2012.Krizhevsky} and VGGNets~\cite{VGGNet.2014.Simonyan}.
However, it is not clear whether those extra features in very deep residual networks can perform as well as conventional fully end-to-end trained features.
So in this paper, we only favour a deeper model, when it can be completely end-to-end trained.

In practice, algorithms are often limited by their spatial costs.
One way is to use more devices, which will however increase communication costs among them.
With similar memory costs, a shallower but wider network can have times more number of trainable parameters.
Therefore, given the following observations in the literature,
\begin{itemize}
\item \vspace{-2.0mm}
Zagoruyko and Komodakis~\cite{WRN.2016.Zagoruyko} found that the performance of a ResNet was related to the number of trainable parameters.
Szegedy~et~al.~\cite{InceptionResNet.2016.Szegedy} came to a similar conclusion, according to the comparison between their proposed Inception networks.
\item \vspace{-2.0mm}
Veit~et~al.~\cite{UnraveledResNet.2016.Veit} found that there is a relatively small effective depth for a very deep ResNet, e.g., seventeen residual units for a 110-layer ResNet.
\vspace{-2.0mm}
\end{itemize}
most of the current state-of-the-art models on the ImageNet classification dataset~\cite{ILSVRC2015.Russakovsky} seem over-deepened, e.g., the 200-layer ResNet~\cite{ResNet2.2016.He} and 96-layer Inception-ResNet~\cite{InceptionResNet.2016.Szegedy}.
The reason is that, to effectively utilize GPU memories, we should make a model shallow.
According to our previous analysis, paths longer than the effective depth in ResNets are not trained in a fully end-to-end manner.
Thus, we can remove most of these paths by directly reducing the number of residual units.
For example, in our best performing network, there are exactly seventeen residual units.

With empirical results, we will show that our fully end-to-end networks can perform much better than the previous much deeper ResNets, especially as feature extractors.
However, even if a rather shallow network (eight-unit, or twenty-layer) can outperform ResNet-152 on the ImageNet classification dataset,
we will not go that shallow, because an appropriate depth is vital to train good features.

\section{Approach to image classification}

\begin{figure}[t]
\begin{center}
\includegraphics[width=0.90\linewidth,trim=0 0 200 0]{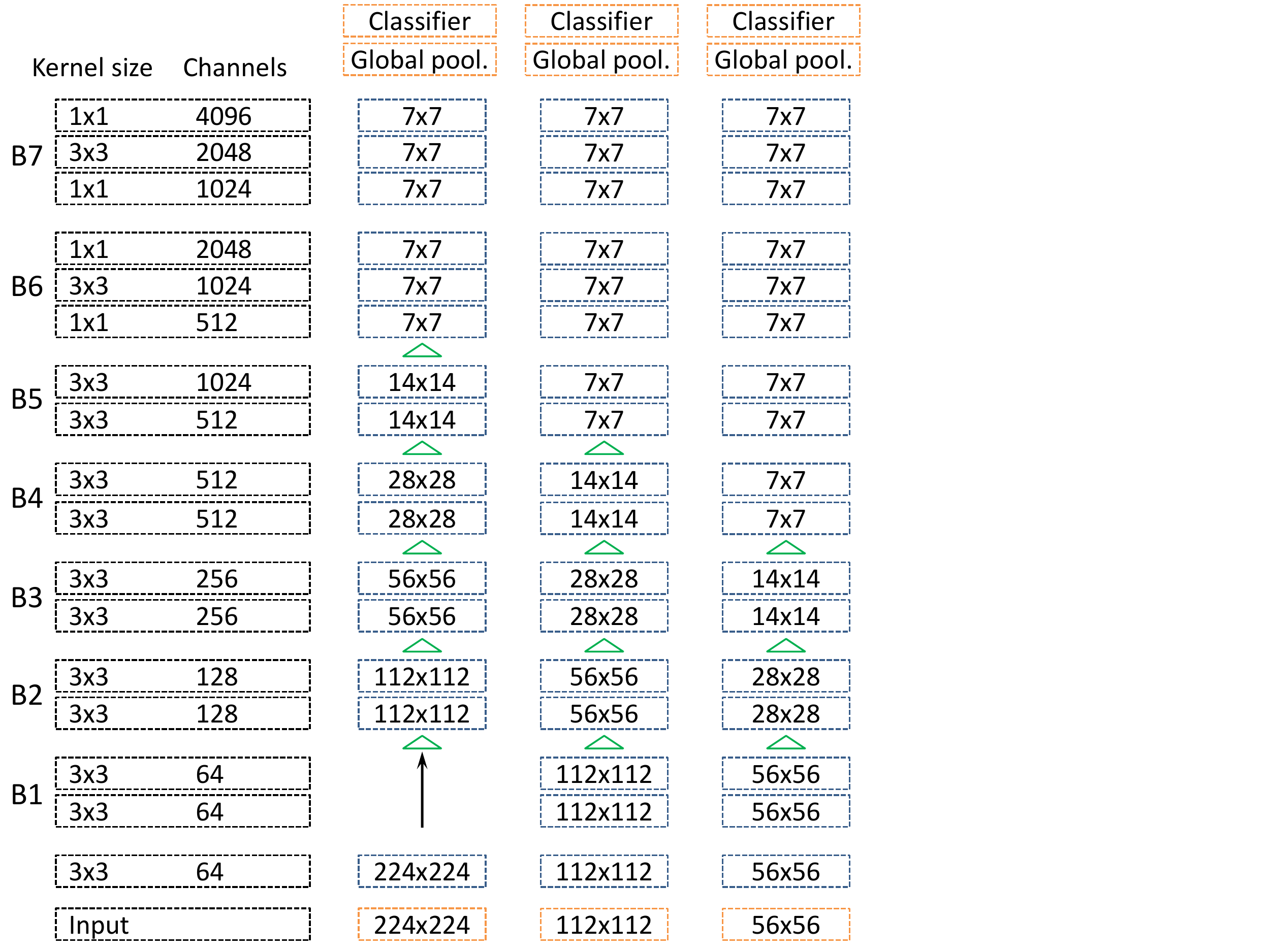}
\end{center}
\caption{Overview of our proposed networks with different input sizes.
Note that B1--B7 are respectively a residual unit.}
\label{fig:structure}
\end{figure}

We show the proposed networks in~Fig.~\ref{fig:structure}.
There are three architectures, with different input sizes.
Dashed blue rectangles to denote convolution stages, which are respectively composed of a batch normalization, an ReLU non-linearity and a convolution layer, following the second version of ResNets~\cite{ResNet2.2016.He}.
The closely stacked two or three convolution stages denote different kinds of residual units~(B1--B7), with inner shortcut connections~\cite{ResNet2.2016.He}.
Each kind corresponds to a level, where all units share the same kernel sizes and numbers of channels, as given in the dashed black rectangles in the left-most column of~Fig.~\ref{fig:structure}.
As mentioned before, there are two 3$\times$3 convolution layers in most residual units~(B1--B5).
However, in B6~and~B7, we use bottleneck structures as in ResNets~\cite{ResNet.CVPR.2016.He},
except that we adjust the numbers of channels to avoid drastic changes in width.
Each of our networks usually consists of one B6, one B7, and different numbers of B1--B5.
For those with a 224$\times$224 input, we do not use B1 due to limited GPU memories.
Each of the green triangles denotes a down-sampling operation with a rate of two, which is clear given the feature map sizes of different convolution stages (in dashed blue rectangles).
To this end, we can let the first convolution layer at according levels have a stride of two.
Or, we can use an extra spatial pooling layer, whose kernel size is three and stride is two.
In a network whose classification results are reported in this paper, we always use pooling layers for down-sampling.
We average the top-most feature maps into 4,096-dimensional final features, which matches the cases of AlexNet~\cite{AlexNet.NIPS.2012.Krizhevsky} and VGGNets~\cite{VGGNet.2014.Simonyan}.
We will show more details about network structures in Subsection~\ref{subsec:imagenet res}.

\textbf{Implementation details}.
We run all experiments using the MXNet framework~\cite{MXNet.2015.Chen}, with four devices (two K80 or four Maxwell Titan X cards) on a single node.
We follow settings in the re-implementation of ResNets by Gross and Wilber~\cite{FacebookResNet.2016.Gross} as possible.
But, we use a linear learning rate schedule, which was reported as a better choice by Mishkin~et~al.~\cite{CaffeNetBenchmark.2016.Mishkin}.
Take Model~A in~Table~\ref{tbl:imagenet} for example.
We start from 0.1, and linearly reduce the learning rate to $10^{-6}$ within 450k iterations.

\section{Approach to semantic image segmentation}

Our approach is similar to the fully convolutional networks (FCN)~\cite{FCN.CVPR.2015.Long} implemented in the first version of DeepLab~\cite{DeepLab.ICLR.2015.Chen}.
However, without getting too many factors entangled, we in this paper do not introduce any multi-scale structures~\cite{AdelaideContext.2016.Lin,DeepLab2.2016.Chen}, deep supervision signals~\cite{FCN.CVPR.2015.Long,SenseCUSceneParsing.2016.Zhao}, or global context features~\cite{SenseCUSceneParsing.2016.Zhao}.
Besides, we do not apply any multi-scale testing, model averaging or CRF based post-processing,
except for the test set of ADE20K~\cite{ADE20K.2016.Zhou}.

Given a pre-trained network, there are three steps to reshape it into a network suitable for semantic image segmentation, as stated below.

1) \textbf{Resolution}.
To generate score maps at 1/8 resolution, we remove down-sampling operations and increase dilation rates accordingly in some convolution layers.
For clarity, first suppose that we always down-sample features maps using a convolution layer with a stride of two.
Take networks with 224$\times$224 inputs for example.
We set stride of the first convolution layer in B5 to one, and increase the dilation rate from one to two for the following layers;
We do the same thing to the first convolution layer in B6 too, and increase the dilation rate from two to four for the following layers.
In the case of down-sampling using a pooling layer,
everything is the same except that we set stride of that pooling layer to one.
Sometimes, we will have to apply a pooling layer with dilation~\cite{DilatedPool.2016.Wu}.
On the other hand, we do not make any change for networks with 56$\times$56 inputs,
since there are only three down-sampling operations in each of them.

It is notable that all down-sampling operations are implemented using spatial pooling layers in our originally pre-trained networks.
We find it harmful for FCNs in our preliminary experiments, probably due to too strong spatial invariance.
To this end, we replace several top-most down-sampling operations in a network,
and then tune it for some additional iterations.
Take Model~A in~Table~\ref{tbl:imagenet} for example again.
We remove the top-most three pooling layers (before B4, B5 and B6),
increase the strides of according convolution layers up to two,
and tune it for 45k iterations using the ImageNet dataset~\cite{ILSVRC2015.Russakovsky}, starting from a learning rate of 0.01.

2) \textbf{Classifier}.
We remove the top-most linear classifier and the global pooling layer,
and then consider two cases.
For one thing, we follow a basic large field-of-view setting in DeepLab-v2~\cite{DeepLab2.2016.Chen}, called~`1 convolution'.
Namely, we just add back a single linear layer as the new classifier.
For anther, we insert an additional non-linear convolution stage (without batch normalization) below the linear classifier.
This case is called~`2 convolutions'.
Both of the added layers have 3$\times$3 kernels, with a dilation rate of twelve.
The top-most two-layer classifier thus has a receptive field of 392$\times$392 on the final feature maps.
By default, we let the number of channels in the hidden layer be 512.

3) \textbf{Dropout}.
To alleviate over-fitting, we also apply the traditional dropout~\cite{Dropout.JMLR.2014.Srivastava} to very wide residual units.
The dropout rate is 0.3 for those with 2,048 channels, e.g.,
the last three units in ResNets and the second last units~(B6) in our networks;
while 0.5 for those with 4,096 channels, e.g.,
the top-most units~(B7) in our networks.

\textbf{Implementation details}.
We fix the moving means and variations in batch normalization layers during fine-tuning~\cite{ResNet.CVPR.2016.He}.
We use four devices on a single node.
The batch size is sixteen, so there are four examples per device.
We first tune each network for a number of iterations, keeping the learning rate unchanged at 0.0016.
And then, we reduce the learning rate gradually during another number of iterations, following a linear schedule~\cite{CaffeNetBenchmark.2016.Mishkin}.
For datasets with available testing sets,
we evaluate these numbers of iterations on validation sets.
During training, we first resize an image by a ratio randomly sampled from $[0.7, 1.3]$,
and then generate a sample by cropping one 500$\times$500 sub-window at a randomly selected location.

\section{Experimental results}

\subsection{Image classification results}
\label{subsec:imagenet res}

We evaluate our proposed networks\footnote{We will release these networks soon.} on the ILSVRC 2012 classification dataset~\cite{ILSVRC2015.Russakovsky}, with 1.28 million images for training, respectively belonging to 1,000 categories.
We report top-1 and top-5 error rates on the validation set.
We compare various networks in Table~\ref{tbl:imagenet},
where we obtain all the results by testing on a single crop.
However, we list the ten-crop result for VGG16~\cite{VGGNet.2014.Simonyan} since it is not inherently a fully convolutional network.
For networks trained with 224$\times$224 inputs,
the testing crop size is 320$\times$320, following the setting used by He~et~al.~\cite{ResNet2.2016.He}.
For those with 112$\times$112 and 56$\times$56 inputs,
we use 160$\times$160 and 80$\times$80 crops respectively.
For Inception networks~\cite{InceptionResNet.2016.Szegedy}, the testing crop size is 299$\times$299~\cite{ResNet2.2016.He}.
The names of our proposed networks are composed of training crop sizes and the numbers of residual units on different levels.
Take 56-1-1-1-1-9-1-1 for example.
Its input size is 56,
and there are only one unit on all levels except for Level~5~(B5 in Fig.~\ref{fig:structure}).

Notable points about the results are as follows.

1) Relatively shallow networks can outperform very deep ones,
which is probably due to large model capacity,
coinciding with the results reported by Zagoruyko and Komodakis~\cite{WRN.2016.Zagoruyko}.
For example, the much shallower Model~B achieves similar error rates as ResNet-152, and even runs slightly faster.
And particularly, Model~A performs the best among all the networks.

2) We can trade performance for efficiency by using a small input size.
For example, Model~D performs slightly worse than ResNet-152, but is almost two times faster.
This may be useful when efficiency is strictly required.
Mishkin~et~al.~\cite{CaffeNetBenchmark.2016.Mishkin} also reduced the input size for efficiency.
However, they did not remove down-sampling operations accordingly to preserve the size of final feature maps,
which resulted in much degraded performance.

3) Models C, D and E perform comparably,
even though Model~C has larger depth and more parameters.
This comparison shows the importance of designing a network properly.
In these models, we put too many layers on low resolution levels~(7$\times$7, B5 in Fig.~\ref{fig:structure}).

\begin{table}
\setlength{\tabcolsep}{2pt}
\small
\begin{center}
\resizebox{0.45\textwidth}{!}{
\begin{tabular}{l|c|c|c|c|c}
\hline
method & depth & tr.~input & top-1 & top-5 & speed \\
\hline\hline
VGG16~\cite{VGGNet.2014.Simonyan}, 10 crops & 16 & 224 & 28.1 & 9.3 & -- \\
ResNet-50~\cite{ResNet.CVPR.2016.He}, our tested & 50 & 224 & 23.5 & 6.8 & 75.2 \\
ResNet-101~\cite{ResNet.CVPR.2016.He}, our tested & 101 & 224 & 22.1 & 6.1 & 56.8 \\
ResNet-152~\cite{ResNet.CVPR.2016.He}, our tested & 152 & 224 & 21.8 & 5.8 & 41.8 \\
ResNet-152~\cite{ResNet2.2016.He} & 152 & 224 & 21.3 & 5.5 & -- \\
ResNet-152~\cite{ResNet2.2016.He}, pre-act. & 152 & 224 & 21.1 & 5.5 & -- \\
ResNet-200~\cite{ResNet2.2016.He}, pre-act. & 200 & 224 & 20.7 & 5.3 & -- \\
Inception-v4~\cite{InceptionResNet.2016.Szegedy} & 76 & 299 & 20.0 & 5.0 & -- \\
Inception-ResNet-v2~\cite{InceptionResNet.2016.Szegedy} & 96 & 299 & 19.9 & 4.9 & -- \\
\hline
56-1-1-1-1-9-1-1, Model~F & 34 & 56 & 25.2 & 7.8 & 113.5 \\
\hline
112-1-1-1-1-5-1-1, Model~E & 26 & 112 & 22.3 & 6.2 & 97.3 \\
112-1-1-1-1-9-1-1, Model~D & 34 & 112 & 22.1 & 6.0 & 81.2 \\
112-1-1-1-1-13-1-1, Model~C & 42 & 112 & 21.8 & 5.9 & 69.2 \\
\hline
224-0-1-1-1-1-1-1 & 16 & 224 & 22.0 & 5.8 & 55.3 \\
224-0-1-1-1-3-1-1, Model~B & 20 & 224 & 21.0 & 5.5 & 43.3 \\
224-0-3-3-6-3-1-1, Model~A & 38 & 224 & \textbf{19.2} & \textbf{4.7} & 15.7 \\
\hline
\end{tabular}
}
\end{center}
\caption{Comparison of networks by top-1 (\%) and top-5 (\%) errors on the ILSVRC 2012 validation set~\cite{ILSVRC2015.Russakovsky} with 50k images, obtained using a single crop.
Testing speeds (images/second) are evaluated with ten~images/mini-batch using cuDNN 4 on a GTX 980 card.
Input sizes during training are also listed.
Note that a smaller size often leads to faster training speed.}
\label{tbl:imagenet}
\end{table}

\subsection{Semantic image segmentation results}

We evaluate our proposed networks on four widely used datasets.
When available, we report,
1) the pixel accuracy, which is the percentage of correctly labelled pixels on a whole test set,
2) the mean pixel accuracy, which is the mean of class-wise pixel accuracies, and
3) the mean IoU score, which is the mean of class-wise intersection-over-union scores.

\textbf{PASCAL VOC 2012}~\cite{PascalVoc.IJCV.2014.Everingham}.
This dataset consists of daily life photos.
There are 1,464 labelled images for training and another 1,449 for validation.
Pixels either belong to the background or twenty object categories, including \textit{bus}, \textit{car}, \textit{cat}, \textit{sofa}, \textit{monitor}, etc.
Following the common criteria in the literature~\cite{FCN.CVPR.2015.Long,DeepLab.ICLR.2015.Chen}, we augment the dataset with extra labelled images from the semantic boundaries dataset~\cite{SBD.ICCV.2011.Hariharan}.
So in total, there are 10,582 images for training.

We first compare different networks in Table~\ref{tbl:voc}.
Notable points about the results are as follows.

1) We cannot make statistically significant improvement by using ResNet-152 instead of ResNet-101.
However, Model~A performs better than ResNet-152 by 3.4\%.
Using one hidden layer leads to a further improvement by 2.1\%.

2) The very deep ResNet-152 uses too many memories due to intentionally enlarged depth. With our settings, it even cannot be tuned using many mainstream GPUs with only 12GB memories.

3) Model~B performs worse than ResNet-101, even if it performs better on the classification task as shown in Table~\ref{tbl:imagenet}.
This shows that it is not reliable to tell a good feature extractor only depending on its classification performance.
And it again shows why we should favour deeper models.

4) Model~A2 performs worse than Model~A on this dataset.
We initialize it using weights from Model~A, and tune it with the Places~365 data~\cite{Places365.2016.Zhou} for 45k iterations.
This is reasonable since there are only object categories in this dataset,
while Places~365 is for scene classification tasks.

\begin{table}
\setlength{\tabcolsep}{3pt}
\small
\begin{center}
\resizebox{0.45\textwidth}{!}{
\begin{tabular}{l|c|c|c|c}
\hline
method & pixel acc. & mean acc. & mean IoU & mem. \\
\hline\hline
ResNet-101, 1 conv. & 94.52 & 82.17 & 75.35 & 12.3 \\
ResNet-152, 1 conv. & 94.56 & 82.12 & 75.37 & {\color{red}{16.7}} \\
Model~F, 1 conv. & 92.91 & 76.26 & 68.36 & 11.7 \\
Model~E, 1 conv. & 94.13 & 80.84 & 73.64 & 10.1 \\
Model~D, 1 conv. & 94.59 & 82.50 & 75.66 & 11.7 \\
Model~C, 1 conv. & 94.56 & 82.01 & 75.45 & 13.4 \\
Model~B, 1 conv. & 93.94 & 80.96 & 73.37 & 7.6 \\
Model~A, 1 conv. & \textbf{95.28} & \textbf{85.23} & \textbf{78.76} & 11.0 \\
Model~A2, 1 conv. & 95.16 & 85.72 & 78.14 & 11.0 \\
\hline
Model~A, 2 conv. & \textbf{95.70} & \textbf{86.26} & \textbf{80.84} & 11.8 \\
\hline
\end{tabular}
}
\end{center}
\caption{Comparison by semantic image segmentation scores~(\%) and GPU memory usages~(GB/device) during tuning on the PASCAL VOC val set~\cite{PascalVoc.IJCV.2014.Everingham} with 1,449 images.
Memory usages are obtained with four images/device using MXNet~\cite{MXNet.2015.Chen}.
}
\label{tbl:voc}
\end{table}

We then compare our method with previous ones on the test set in~Table~\ref{tbl:voc test}.
Only using the augmented PASCAL VOC data for training,
we achieve a mean IoU score of 82.5\%\footnote{
\resizebox{0.45\textwidth}{!}{
\url{http://host.robots.ox.ac.uk:8080/anonymous/H0KLZK.html}
}
},
which is better than the previous best one by 3.4\%.
This is a significant margin, considering that the gap between ResNet-based and VGGNet-based methods is 3.8\%.
Our method wins for seventeen out of the twenty object categories, which was the official criteria used in the PASCAL VOC challenges~\cite{PascalVoc.IJCV.2014.Everingham}.
In some works~\cite{AdelaideContext.2016.Lin,DeepLab2.2016.Chen},
models were further pre-trained using the Microsoft COCO~\cite{COCO.ECCV.2014.Lin} data,
which consists of 120k labelled images.
In this case, the current best mean IoU is 79.7\% reported by Chen~et~al.~\cite{DeepLab2.2016.Chen}.
They also used multi-scale structure and CRF-based post-processing in their submission, which we do not consider here.
Nevertheless, our method outperforms theirs by 2.8\%, which further shows the effectiveness of our features pre-trained only using the ImageNet classification data~\cite{ILSVRC2015.Russakovsky}.

\begin{table*}[!ht]
\setlength{\tabcolsep}{2pt}
\small
\begin{center}
\resizebox{0.95\textwidth}{!}{
\begin{tabular}{l|cccccccccccccccccccc|c}
\toprule
method & \rot{aero.} & \rot{bicy.} & \rot{bird} & \rot{boat} & \rot{bott.} & \rot{bus} & \rot{car} & \rot{cat} & \rot{chai.} & \rot{cow} & \rot{dini.} & \rot{dog} & \rot{hors.} & \rot{moto.} & \rot{pers.} & \rot{pott.} & \rot{shee.} & \rot{sofa} & \rot{trai.} & \rot{tvmo.} & \rot{mean} \\
\hline\hline
\multicolumn{22}{c}{using augmented PASCAL VOC data only} \\
\hline
FCN-8s~\cite{FCN.CVPR.2015.Long} & 76.8 & 34.2 & 68.9 & 49.4 & 60.3 & 75.3 & 74.7 & 77.6 & 21.4 & 62.5 & 46.8 & 71.8 & 63.9 & 76.5 & 73.9 & 45.2 & 72.4 & 37.4 & 70.9 & 55.1 & 62.2 \\
CRFasRNN~\cite{CRFasRNN.ICCV.2015.Zheng} & 87.5 & 39.0 & 79.7 & 64.2 & 68.3 & 87.6 & 80.8 & 84.4 & 30.4 & 78.2 & 60.4 & 80.5 & 77.8 & 83.1 & 80.6 & 59.5 & 82.8 & 47.8 & 78.3 & 67.1 & 72.0 \\
DeconvNet~\cite{DeconvNet.ICCV.2015.Noh} & 89.9 & 39.3 & 79.7 & 63.9 & 68.2 & 87.4 & 81.2 & 86.1 & 28.5 & 77.0 & 62.0 & 79.0 & 80.3 & 83.6 & 80.2 & 58.8 & 83.4 & 54.3 & 80.7 & 65.0 & 72.5 \\
DPN~\cite{DPN.ICCV.2015.Liu} & 87.7 & 59.4 & 78.4 & 64.9 & 70.3 & 89.3 & 83.5 & 86.1 & 31.7 & 79.9 & 62.6 & 81.9 & 80.0 & 83.5 & 82.3 & 60.5 & 83.2 & 53.4 & 77.9 & 65.0 & 74.1 \\
Context~\cite{AdelaideContext.2016.Lin} & 90.6 & 37.6 & 80.0 & 67.8 & 74.4 & 92.0 & 85.2 & 86.2 & 39.1 & 81.2 & 58.9 & 83.8 & 83.9 & 84.3 & 84.8 & 62.1 & 83.2 & 58.2 & 80.8 & 72.3 & 75.3 \\
VeryDeep$^*$~\cite{InstanceSegmentation.2016.Wu} & 91.9 & 48.1 & 93.4 & \textbf{69.3} & 75.5 & \textbf{94.2} & 87.5 & \textbf{92.8} & 36.7 & 86.9 & 65.2 & 89.1 & 90.2 & 86.5 & 87.2 & 64.6 & 90.1 & 59.7 & 85.5 & 72.7 & 79.1 \\
\hline
Model A, 2 conv. & \textbf{94.4} & \textbf{72.9} & \textbf{94.9} & 68.8 & \textbf{78.4} & 90.6 & \textbf{90.0} & 92.1 & \textbf{40.1} & \textbf{90.4} & \textbf{71.7} & \textbf{89.9} & \textbf{93.7} & \textbf{91.0} & \textbf{89.1} & \textbf{71.3} & \textbf{90.7} & \textbf{61.3} & \textbf{87.7} & \textbf{78.1} & \textbf{82.5} \\
\hline\hline
\multicolumn{22}{c}{using augmented PASCAL VOC \& COCO data} \\
\hline
Context~\cite{AdelaideContext.2016.Lin} & 94.1 & 40.4 & 83.6 & 67.3 & 75.6 & 93.4 & 84.4 & 88.7 & 41.6 & 86.4 & 63.3 & 85.5 & 89.3 & 85.6 & 86.0 & 67.4 & 90.1 & 62.6 & 80.9 & 72.5 & 77.8 \\
DeepLab-v2$^*$~\cite{DeepLab2.2016.Chen} & 92.6 & 60.4 & 91.6 & 63.4 & 76.3 & 95.0 & 88.4 & 92.6 & 32.7 & 88.5 & 67.6 & 89.6 & 92.1 & 87.0 & 87.4 & 63.3 & 88.3 & 60.0 & 86.8 & 74.5 & 79.7 \\
\bottomrule
\end{tabular}
}
\end{center}
\caption{Comparison with previous results by mean intersection-over-union scores~(\%) on the PASCAL VOC test set~\cite{PascalVoc.IJCV.2014.Everingham} with 1,456 images.
Asterisked methods use ResNet-101~\cite{ResNet.CVPR.2016.He},
while others use VGG16~\cite{VGGNet.2014.Simonyan}.}
\label{tbl:voc test}
\end{table*}

\textbf{Cityscapes}~\cite{Cityscapes.CVPR.2016.Cordts}.
This dataset consists of street scene photos taken by car-carried cameras.
There are 2975 labelled images for training and another 500 for validation.
Besides, there is also an extended set with 19,998 coarsely labelled images.
Pixels belong to nineteen semantic classes, including \textit{road}, \textit{car}, \textit{pedestrian}, \textit{bicycle}, etc.
These classes further belong to seven categories, i.e., \textit{flat}, \textit{nature}, \textit{object}, \textit{sky}, \textit{construction}, \textit{human}, and \textit{vehicle}.

We first compare different networks in Table~\ref{tbl:cityscapes}.
On this dataset, ResNet-152 again shows no advantage against ResNet-101.
However, Model~A1 outperforms ResNet-101 by 4.2\% in terms of mean IoU scores, which again is a significant margin.
Because there are many scene classes,
models pre-trained using Places~365~\cite{Places365.2016.Zhou} are supposed to perform better,
which coincides with our results.

\begin{table}
\setlength{\tabcolsep}{7pt}
\small
\begin{center}
\resizebox{0.45\textwidth}{!}{
\begin{tabular}{l|c|c|c}
\hline
method & pixel acc. & mean acc. & mean IoU \\
\hline\hline
\multicolumn{4}{c}{results on the Cityscapes val set} \\
\hline
ResNet-101, 1 conv. & 95.49 & 81.76 & 73.63 \\
ResNet-152, 1 conv. & 95.53 & 81.61 & 73.50 \\
Model~A, 1 conv. & 95.80 & 83.81 & 76.57 \\
Model~A2, 1 conv. & \textbf{95.91} & \textbf{84.48} & \textbf{77.18} \\
\hline
Model~A2, 2 conv. & \textbf{96.05} & \textbf{84.96} & \textbf{77.86} \\
\hline\hline
\multicolumn{4}{c}{results on the ADE20K val set} \\
\hline
ResNet-101, 2 conv. & 79.07 & 48.73 & 39.40 \\
ResNet-152, 2 conv. & 79.33 & 49.55 & 39.77 \\
Model~E, 2 conv. & 79.61 & 50.46 & 41.00 \\
Model~D, 2 conv. & 79.87 & 51.34 & 41.91 \\
Model~C, 2 conv. & 80.53 & 52.32 & 43.06 \\
Model~A, 2 conv. & 80.41 & 52.86 & 42.71 \\
Model~A2, 2 conv. & \textbf{81.17} & \textbf{53.84} & \textbf{43.73} \\
\hline
\end{tabular}
}
\end{center}
\caption{Comparison by semantic image segmentation scores~(\%) on the Cityscapes val set~\cite{Cityscapes.CVPR.2016.Cordts} with 500 images, and the ADE20K val set~\cite{ADE20K.2016.Zhou} with 2k images.}
\label{tbl:cityscapes}
\label{tbl:ade20k}
\end{table}

We then compare our method with previous ones on the test set in~Table~\ref{tbl:cityscapes test}.
The official criteria on this dataset includes two levels, i.e., \textit{class} and \textit{category}.
Besides, there is also an instance-weighted IoU score for each of the two,
which assigns high scores to those pixels of small instances.
Namely, this score penalizes methods ignoring small instances,
which may cause fatal problems in vehicle-centric scenarios.
Our method achieves a class-level IoU score of 78.4\%\footnote{
\resizebox{0.475\textwidth}{!}{
\url{https://www.cityscapes-dataset.com/benchmarks}
}
}, and outperforms the previous best one by 6.6\%.
Furthermore, in the case of instance-weighted IoU score, our method also performs better than the previous best one by 6.4\%.
It is notable that these significant improvements show the strength of our pre-trained features,
considering that DeepLab-v2~\cite{DeepLab2.2016.Chen} uses ResNet-101,
and LRR~\cite{LRR.2016.Ghiasi} uses much more data for training.

\begin{table}
\setlength{\tabcolsep}{5pt}
\small
\begin{center}
\resizebox{0.45\textwidth}{!}{
\begin{tabular}{l|c|c|c|c}
\hline
method & cla.~IoU & cla.~iIoU & cat.~IoU & cat.~iIoU \\
\hline\hline
Dilation10~\cite{Dilation10.2016.Yu} & 67.1 & 42.0 & 86.5 & 71.1 \\
DeepLab-v2$^*$~\cite{ResNet2.2016.He} & 70.4 & 42.6 & 86.4 & 67.7 \\
Context~\cite{AdelaideContext.2016.Lin} & 71.6 & 51.7 & 87.3 & 74.1 \\
LRR$^+$~\cite{LRR.2016.Ghiasi} & 71.8 & 47.9 & 88.4 & 73.9 \\
\hline
Model~A2, 2 conv. & \textbf{78.4} & \textbf{59.1} & \textbf{90.9} & \textbf{81.1} \\
\hline
\end{tabular}
}
\end{center}
\caption{Comparison by semantic image segmentation scores~(\%) on the Cityscapes test set~\cite{Cityscapes.CVPR.2016.Cordts} with 1,525 images.
DeepLab-v2~\cite{ResNet2.2016.He} uses ResNet-101~\cite{ResNet.CVPR.2016.He},
while others use VGG16~\cite{VGGNet.2014.Simonyan}.
LRR~\cite{LRR.2016.Ghiasi} also uses the coarse set for training.}
\label{tbl:cityscapes test}
\end{table}

\textbf{ADE20K}~\cite{ADE20K.2016.Zhou}.
This dataset consists of both indoor and outdoor images with large variations.
There are 20,210 labelled images for training and another 2k for validation.
Pixels belong to 150 semantic categories, including \textit{sky}, \textit{house}, \textit{bottle}, \textit{food}, \textit{toy}, etc.

We first compare different networks in Table~\ref{tbl:ade20k}.
On this dataset, ResNet-152 performs slightly better than ResNet-101.
However, Model~A2 outperforms ResNet-152 by 4.0\% in terms of mean IoU scores.
Being similar with Cityscapes, this dataset has many scene categories.
So, Model~A2 performs slightly better than Model~A.
Another notable point is that, Model~C takes the second place on this dataset,
even if it performs worse than Model~A in the image classification task on the ImageNet dataset.
This shows that large model capacity may become more critical in complicated tasks,
since there are more parameters in Model~C.

We then compare our method with others on the test set in~Table~\ref{tbl:ade20k test}.
The official criteria on this dataset is the average of pixel accuracies and mean IoU scores.
For better performance, we apply multi-scale testing, model averaging and post-processing with CRFs.
Our Model~A2 performs the best among all methods using only a single pre-trained model.
However, in this submission, we only managed to include two kinds of pre-trained features, i.e., Models~A~and~C.
Nevertheless, our method only performs slightly worse than the winner by a margin of 0.47\%.

\begin{table}
\setlength{\tabcolsep}{2pt}
\small
\begin{center}
\resizebox{0.45\textwidth}{!}{
\begin{tabular}{l|c|c}
\hline
\multicolumn{2}{l|}{method} & ave.~of pixel acc.~\& mean IoU \\
\hline
\multicolumn{2}{l|}{SegModel} & 53.23 \\
\multicolumn{2}{l|}{CASIA\_IVA} & 54.33 \\
\multicolumn{2}{l|}{360+MCG-ICT-CAS\_SP} & 54.68 \\
\multicolumn{2}{l|}{SenseCUSceneParsing} & 55.38 \\
\hline
\multicolumn{2}{l|}{ours} & \textbf{56.41} \\
\hline\hline
method & models & ave.~of pixel acc.~\& mean IoU \\
\hline
NTU-SP & 2 & 53.57 \\
SegModel & 5 & 54.65 \\
360+MCG-ICT-CAS\_SP & -- & 55.57 \\
SenseCUSceneParsing & -- & \textbf{57.21} \\
\hline
ours & 2 & 56.74 \\
\hline
\end{tabular}
}
\end{center}
\caption{Comparison by semantic image segmentation scores~(\%) on the ADE20K test set~\cite{ADE20K.2016.Zhou} with 3,352 images.}
\label{tbl:ade20k test}
\end{table}

\textbf{PASCAL Context}~\cite{PascalContext.CVPR.2014.Mottaghi}.
This dataset consists of images from PASCAL VOC 2010~\cite{PascalVoc.IJCV.2014.Everingham} with extra object and stuff labels.
There are 4,998 images for training and another 5,105 for validation.
Pixels either belong to the background category or 59 semantic categories, including \textit{bag}, \textit{food}, \textit{sign}, \textit{ceiling}, \textit{ground} and \textit{snow}.
All images in this dataset are no larger than 500$\times$500.
Since the test set is not available, here we directly apply the hyper-parameters which are used on the PASCAL VOC dataset.
Our method again performs the best with a clear margin by all the three kinds of scores, as shown in~Table~\ref{tbl:pascal-context}.
In particular, we improve the IoU score by 2.4\% compared to the previous best method~\cite{DeepLab2.2016.Chen}.

\begin{table}
\setlength{\tabcolsep}{3pt}
\small
\begin{center}
\resizebox{0.45\textwidth}{!}{
\begin{tabular}{l|l|c|c|c}
\hline
method & feature & pixel acc. & mean acc. & mean IoU \\
\hline\hline
FCN-8s~\cite{FCN.CVPR.2015.Long} & VGG16 & 65.9 & 46.5 & 35.1 \\
BoxSup~\cite{BoxSup.ICCV.2015.Dai} & VGG16 & -- & -- & 40.5 \\
Context~\cite{AdelaideContext.2016.Lin} & VGG16 & 71.5 & 53.9 & 43.3 \\
VeryDeep~\cite{InstanceSegmentation.2016.Wu} & ResNet-101 & 72.9 & 54.8 & 44.5 \\
DeepLab-v2~\cite{DeepLab2.2016.Chen} & ResNet-101 & -- & -- & 45.7 \\
\hline
\multicolumn{2}{l|}{Model A2, 2 conv.} & \textbf{75.0} & \textbf{58.1} & \textbf{48.1} \\
\hline
\end{tabular}
}
\end{center}
\caption{Comparison by semantic image segmentation scores~(\%) on the PASCAL Context val set~\cite{PascalContext.CVPR.2014.Mottaghi} with 5,105 images.}
\label{tbl:pascal-context}
\end{table}

\section{Conclusion}

We have analysed the ResNet architecture, in terms of the ensemble classifiers therein and the effective depths of the residual units.
On the basis of that analysis we calculated a new, more spatially efficient, and better performing architecture which actually achieves fully end-to-end training for large networks.
Using this new architecture we designed a group of correspondingly shallow networks, and showed that they outperform the previous very deep residual networks not only on the ImageNet classification dataset, but also when applied to semantic image segmentation.
These results show that the proposed architecture delivers better feature extraction performance than the current state-of-the-art.

\include{appendix}

{
\bibliographystyle{ieee}
\bibliography{ref}
}

\end{document}

%% file: appendix.tex
\appendix

\onecolumn

\section{Appendix}

\subsection{Network structures}
The graph structures of  Model~A for the ImageNet (ILSVRC 2012)~\cite{ILSVRC2015.Russakovsky} classification 
can be accessed at:

\url{https://cdn.rawgit.com/itijyou/ademxapp/master/misc/ilsvrc_model_a.pdf}

 Model~A2 for the \mbox{PASCAL~VOC~2012~\cite{PascalVoc.IJCV.2014.Everingham}} segmentation %
can be accessed at:

\url{https://cdn.rawgit.com/itijyou/ademxapp/master/misc/voc_model_a2.pdf}.

\subsection{Gradients in residual networks}

\begin{figure*}[h]
\begin{center}
\includegraphics[width=0.45\linewidth,trim=0 0 0 0]{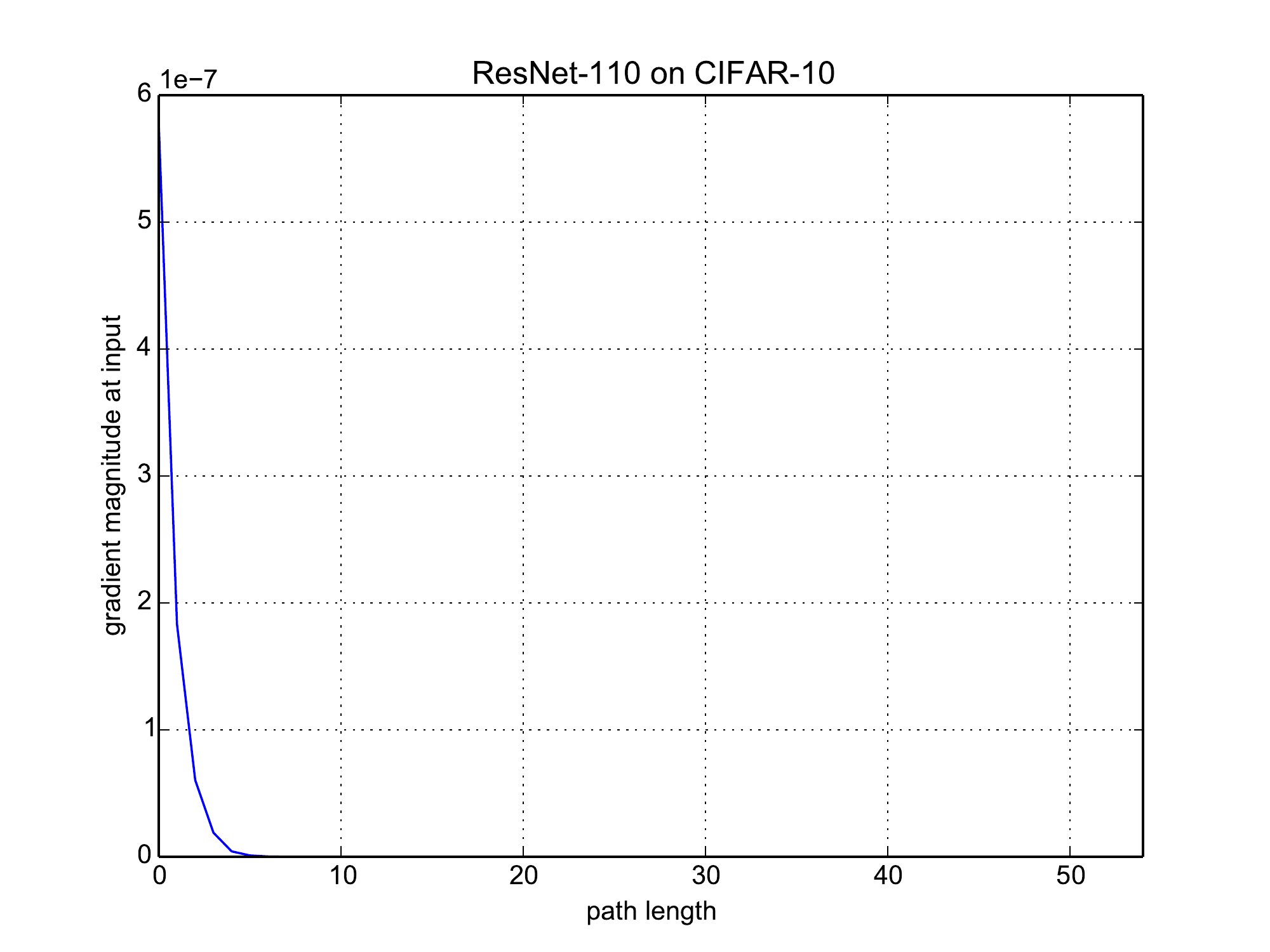}
\includegraphics[width=0.45\linewidth,trim=0 0 0 0]{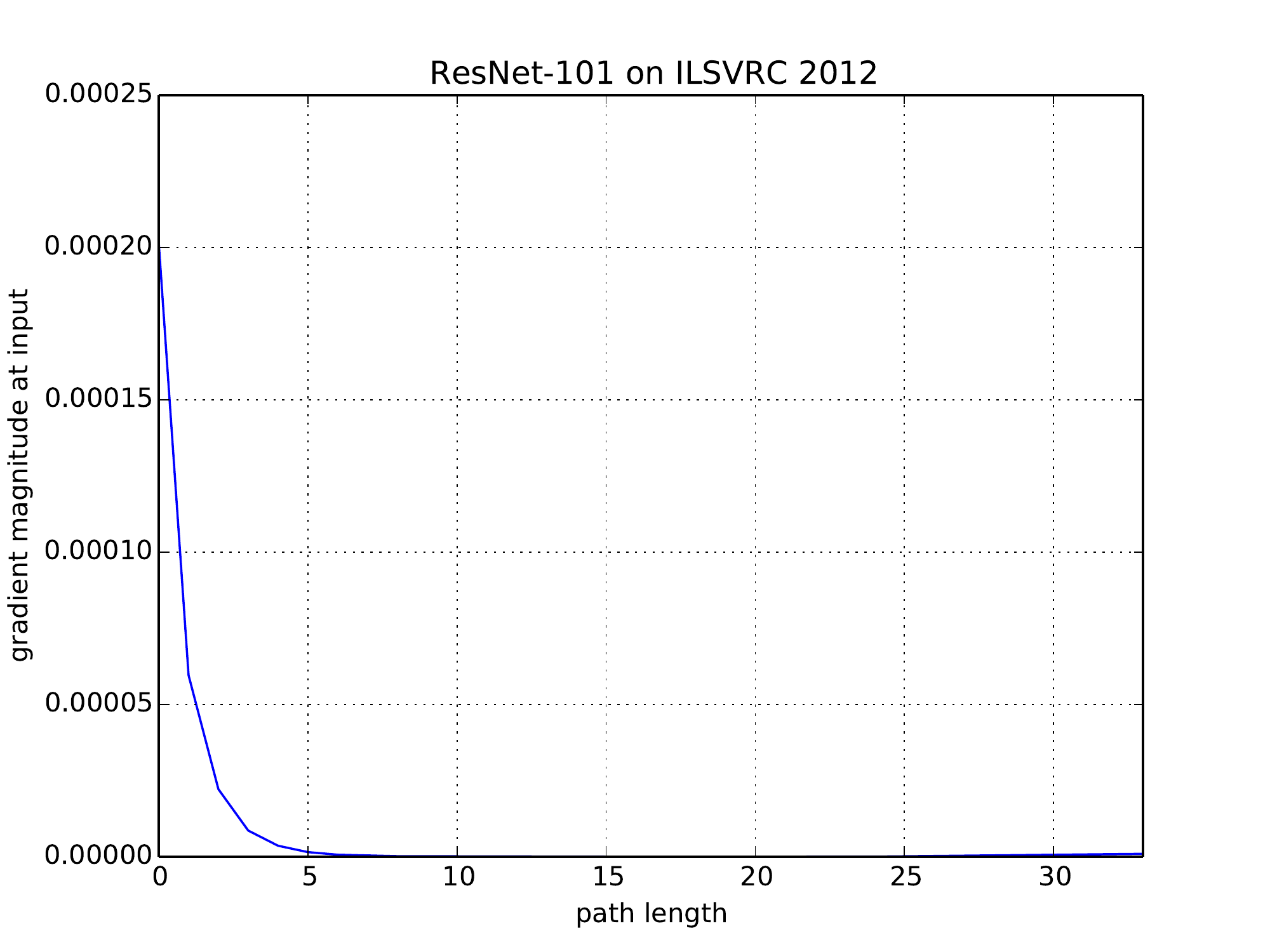}
\includegraphics[width=0.45\linewidth,trim=0 0 0 0]{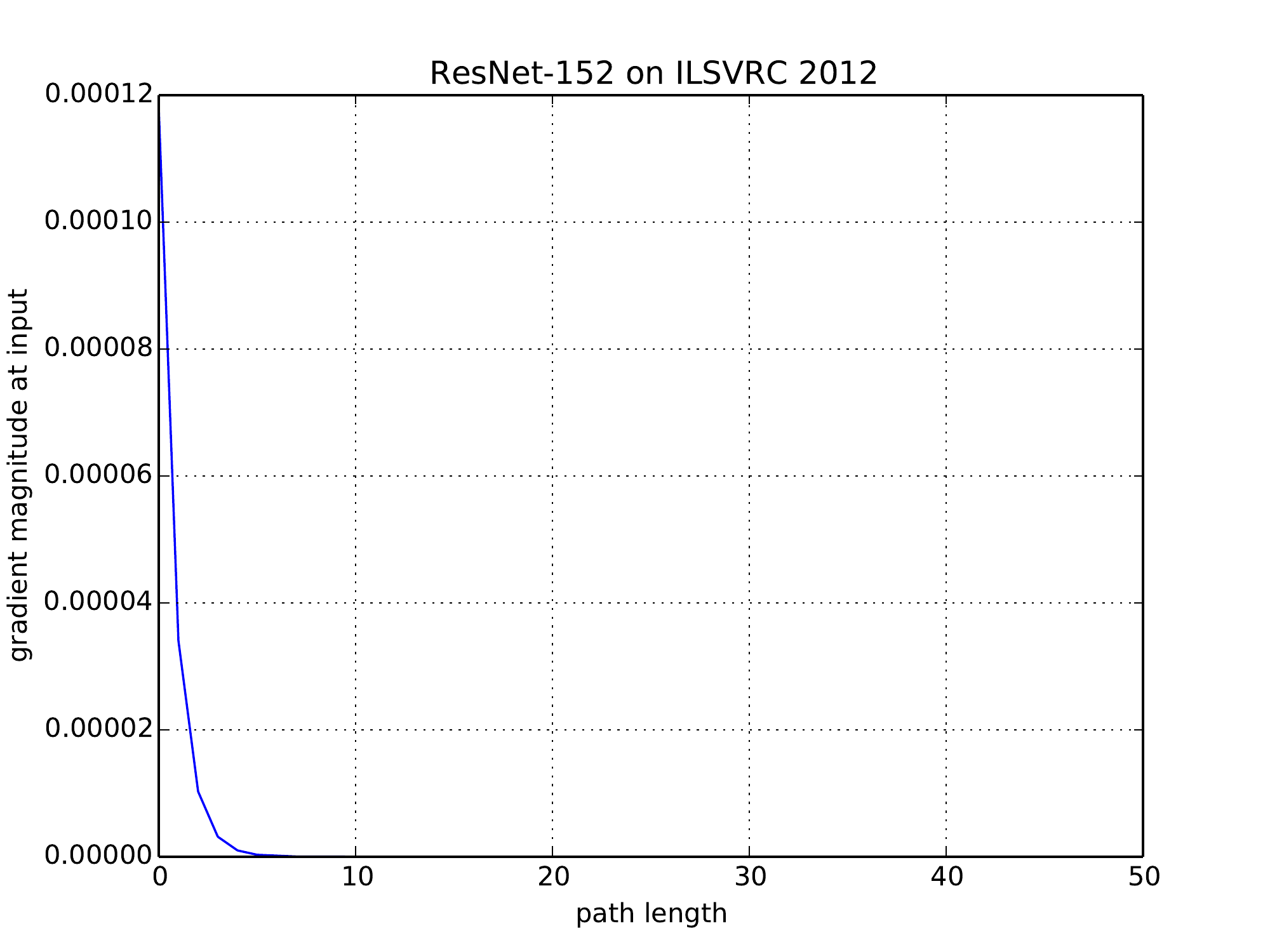}
\includegraphics[width=0.45\linewidth,trim=0 0 0 0]{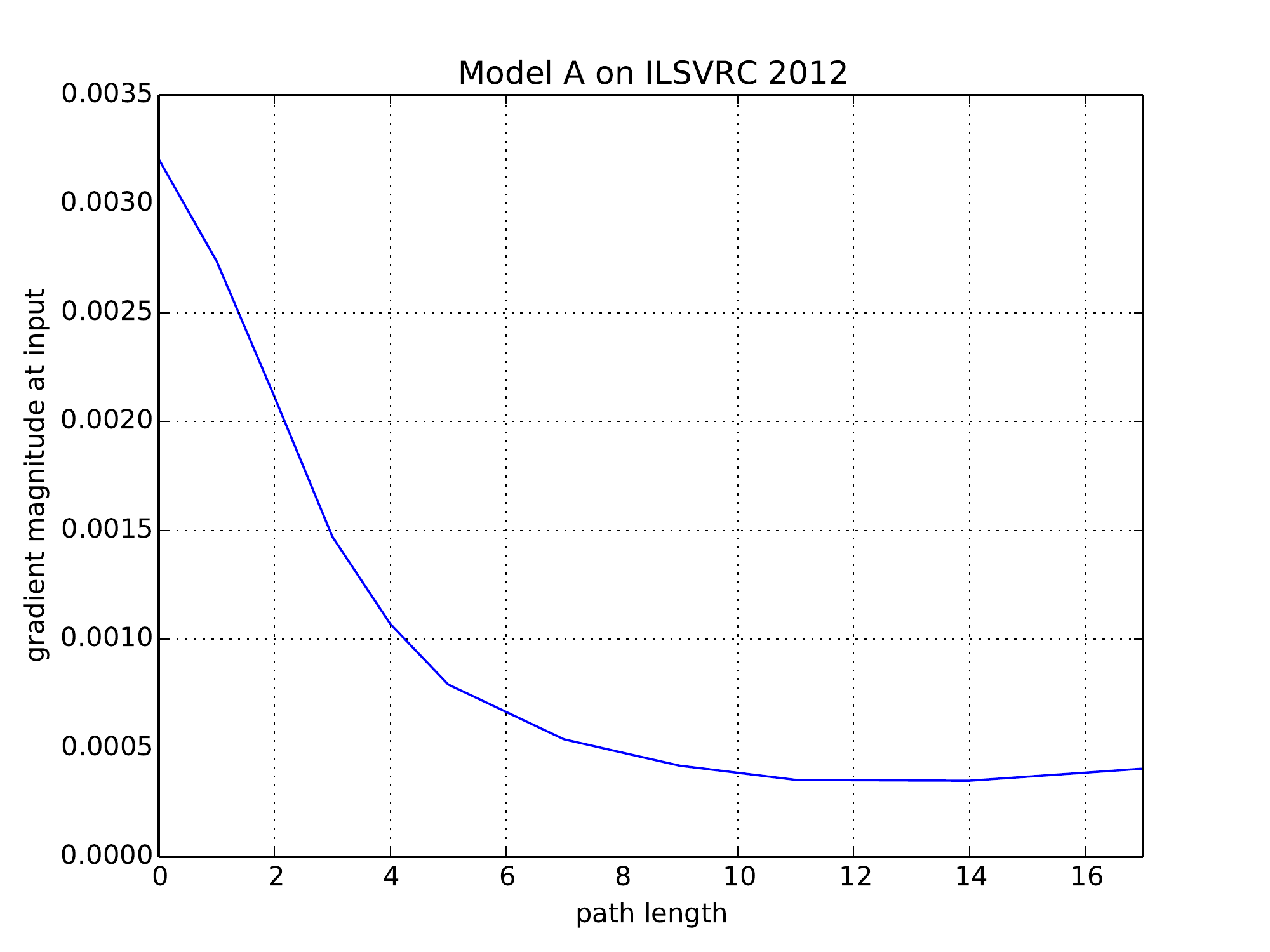}
\end{center}
\caption{
Gradient magnitude at input given a path length $k$ in various residual networks.
See the text for details.
}
\label{fig:gradients}
\end{figure*}

We show results of the experiment on gradients proposed by Veit~et~al.~\cite{UnraveledResNet.2016.Veit}, with various residual networks.
Namely, for a trained network with $n$ units,
we sample individual paths of a certain length $k$,
and measure the norm of gradients that arrive at the input.
Each time, we first feed a batch forward through the whole network;
then during the backward pass,
we randomly sample $k$ units.
For them, we only propagate gradients through their trainable mapping functions, but without their shortcut connections.
For the remaining $n - k$ units, we do the opposite, namely, only propagating gradients through their shortcut connections.
We record the norm of those gradients that reach the input for varying path length $k$,
and show the results in~Fig.~\ref{fig:gradients}.
Note the varying magnitude and maximum path length in individual figures.
These are compared to the middle part of Fig.~6~in~\cite{UnraveledResNet.2016.Veit}.
However, differently we further divide the computed norm of a batch by its number of examples.
According to the results in~Fig.~\ref{fig:gradients},
ResNet-110 trained on CIFAR-10, as well as ResNet-101 and ResNet-152 trained on ILSVRC 2012,
generate much smaller gradients from their long paths than from their short paths.
In contrast, our Model~A trained on ILSVRC 2012, generates more comparable gradients from its paths with different lengths.

\subsection{Qualitative results}
\begin{figure*}[h]
\begin{center}
\includegraphics[width=0.15\linewidth,trim=0 0 0 0]{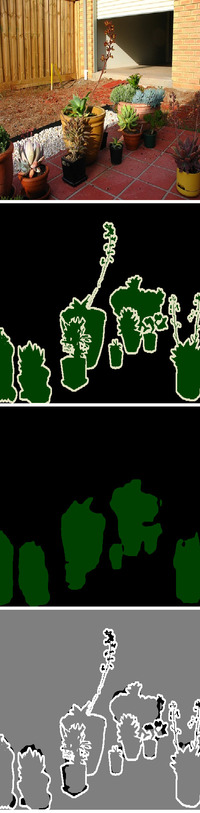}
\includegraphics[width=0.15\linewidth,trim=0 0 0 0]{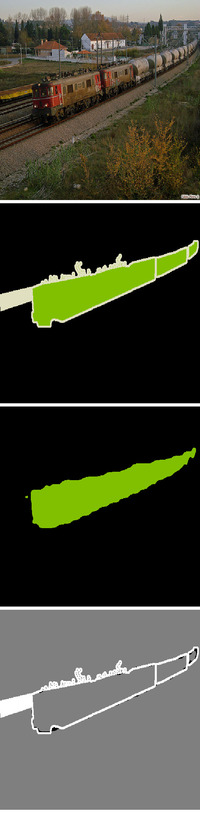}
\includegraphics[width=0.15\linewidth,trim=0 0 0 0]{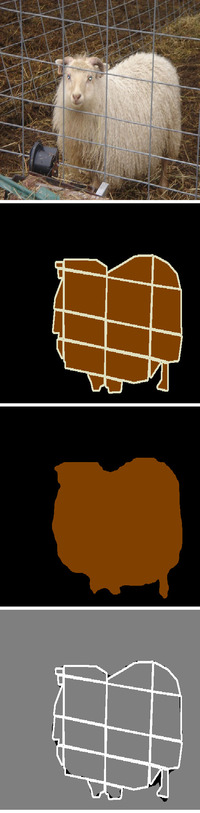}
\includegraphics[width=0.15\linewidth,trim=0 0 0 0]{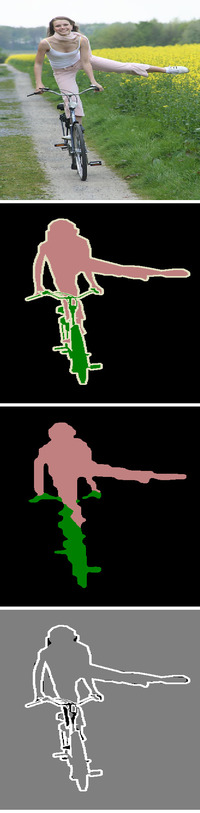}
\includegraphics[width=0.15\linewidth,trim=0 0 0 0]{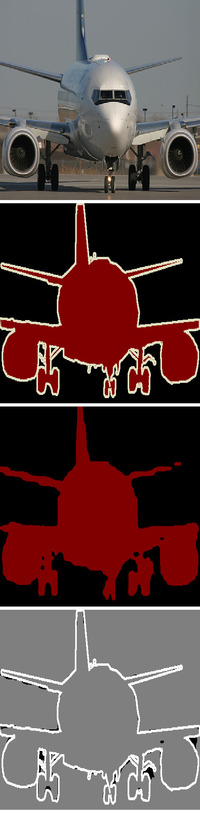}
\includegraphics[width=0.15\linewidth,trim=0 0 0 0]{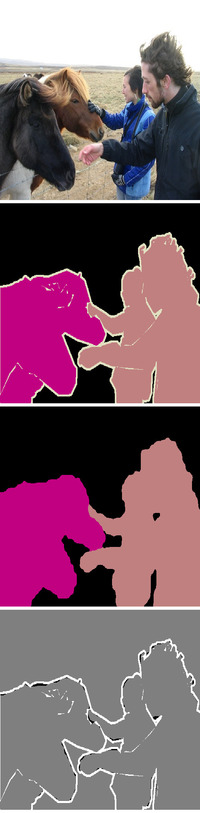}
\includegraphics[width=0.15\linewidth,trim=0 0 0 0]{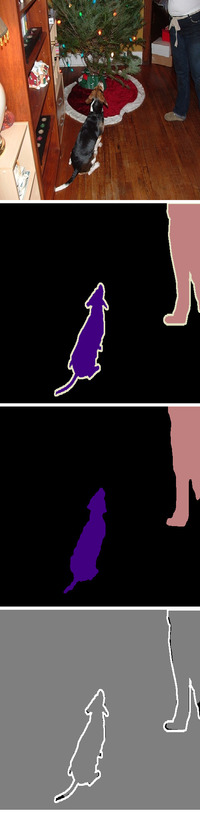}
\includegraphics[width=0.15\linewidth,trim=0 0 0 0]{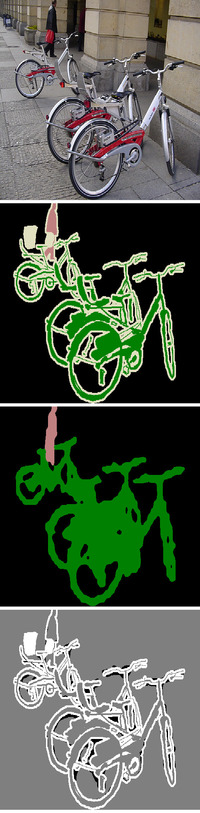}
\includegraphics[width=0.15\linewidth,trim=0 0 0 0]{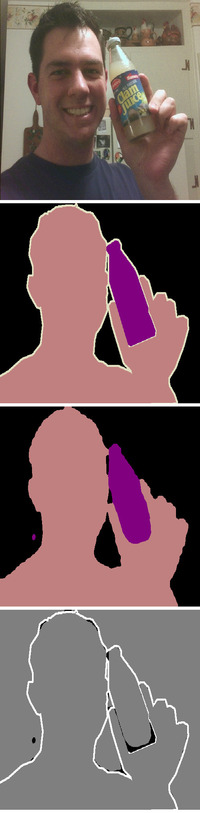}
\includegraphics[width=0.15\linewidth,trim=0 0 0 0]{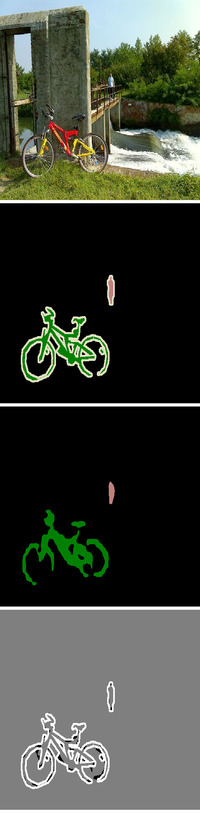}
\includegraphics[width=0.15\linewidth,trim=0 0 0 0]{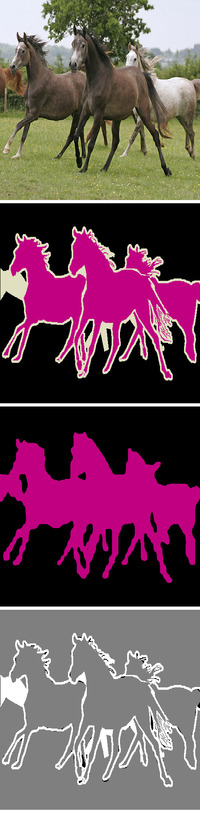}
\includegraphics[width=0.15\linewidth,trim=0 0 0 0]{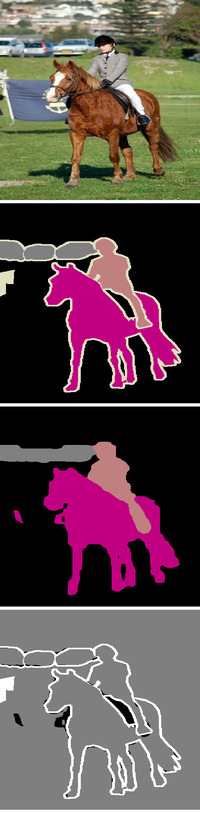}
\end{center}
\caption{
Qualitative results on the PASCAL VOC 2012~\cite{PascalVoc.IJCV.2014.Everingham} val set.
The model was trained using the train set augmented using SBD~\cite{SBD.ICCV.2011.Hariharan}.
In each example, from top to bottom, there are in turn the original image, the ground-truth, the predicted label, and the difference map between the ground-truth and the predicted label.
}
\label{fig:voc}
\end{figure*}

\begin{figure*}[h]
\begin{center}
\includegraphics[width=0.30\linewidth,trim=0 0 0 0]{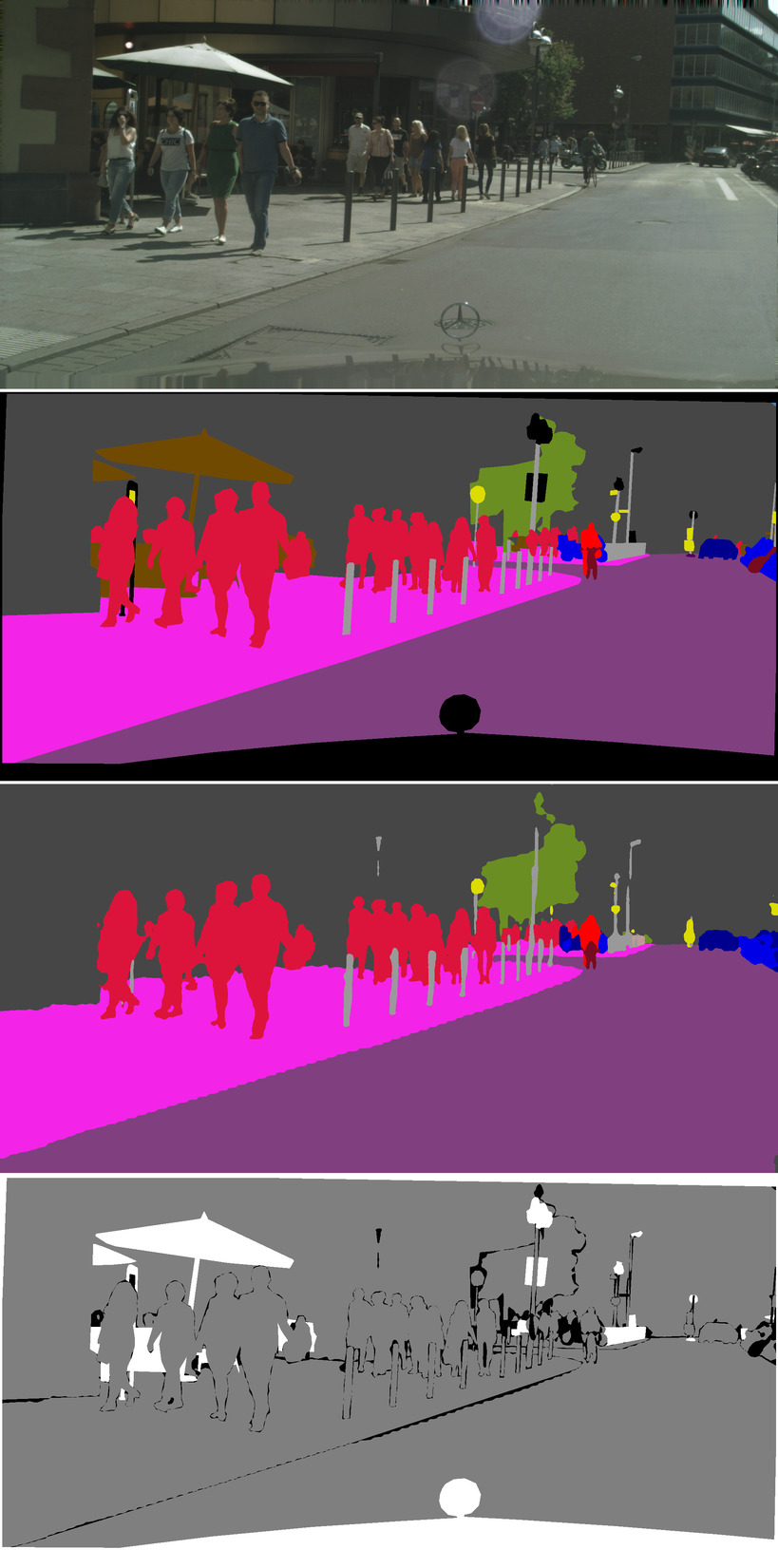}
\includegraphics[width=0.30\linewidth,trim=0 0 0 0]{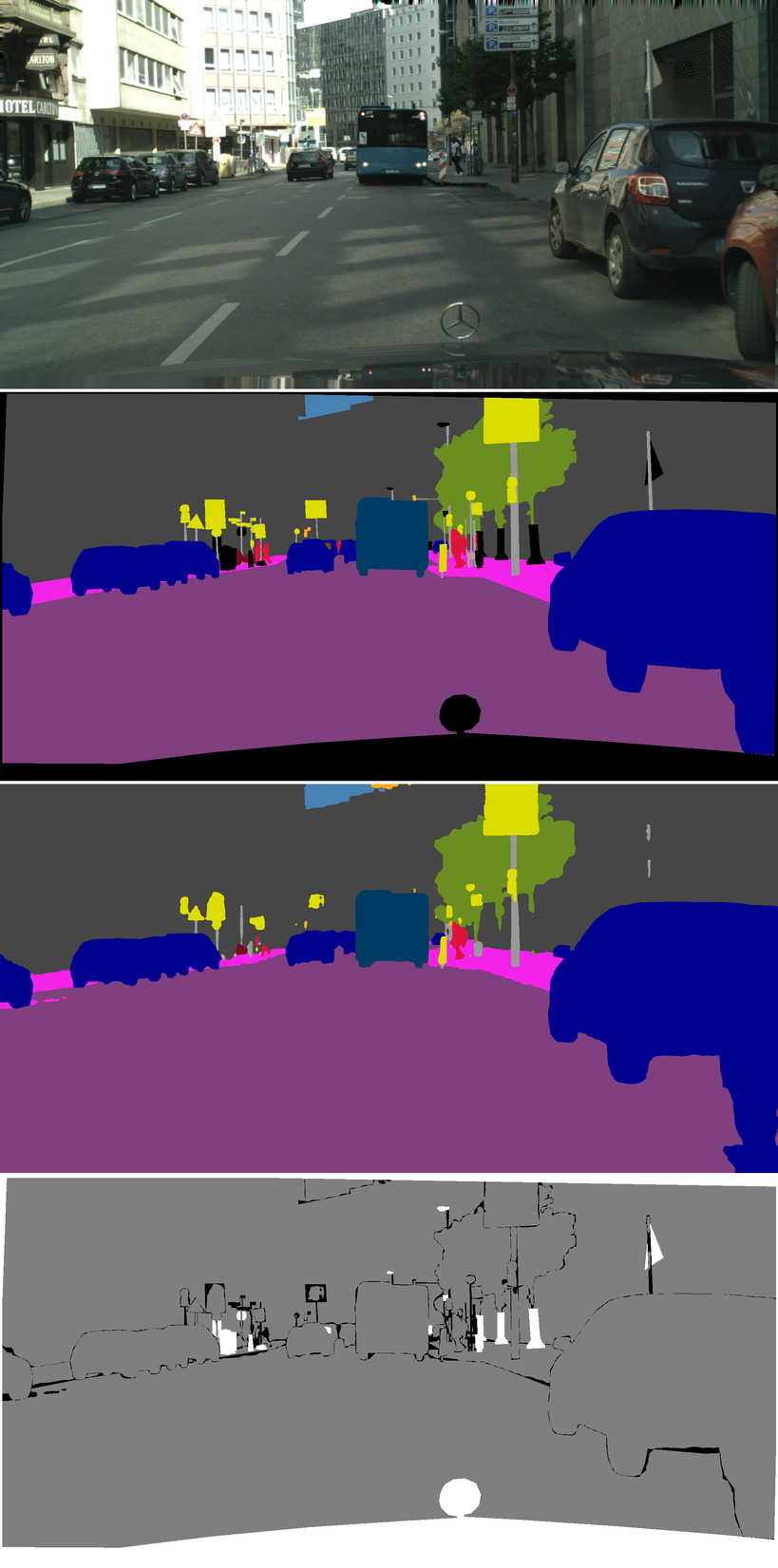}
\includegraphics[width=0.30\linewidth,trim=0 0 0 0]{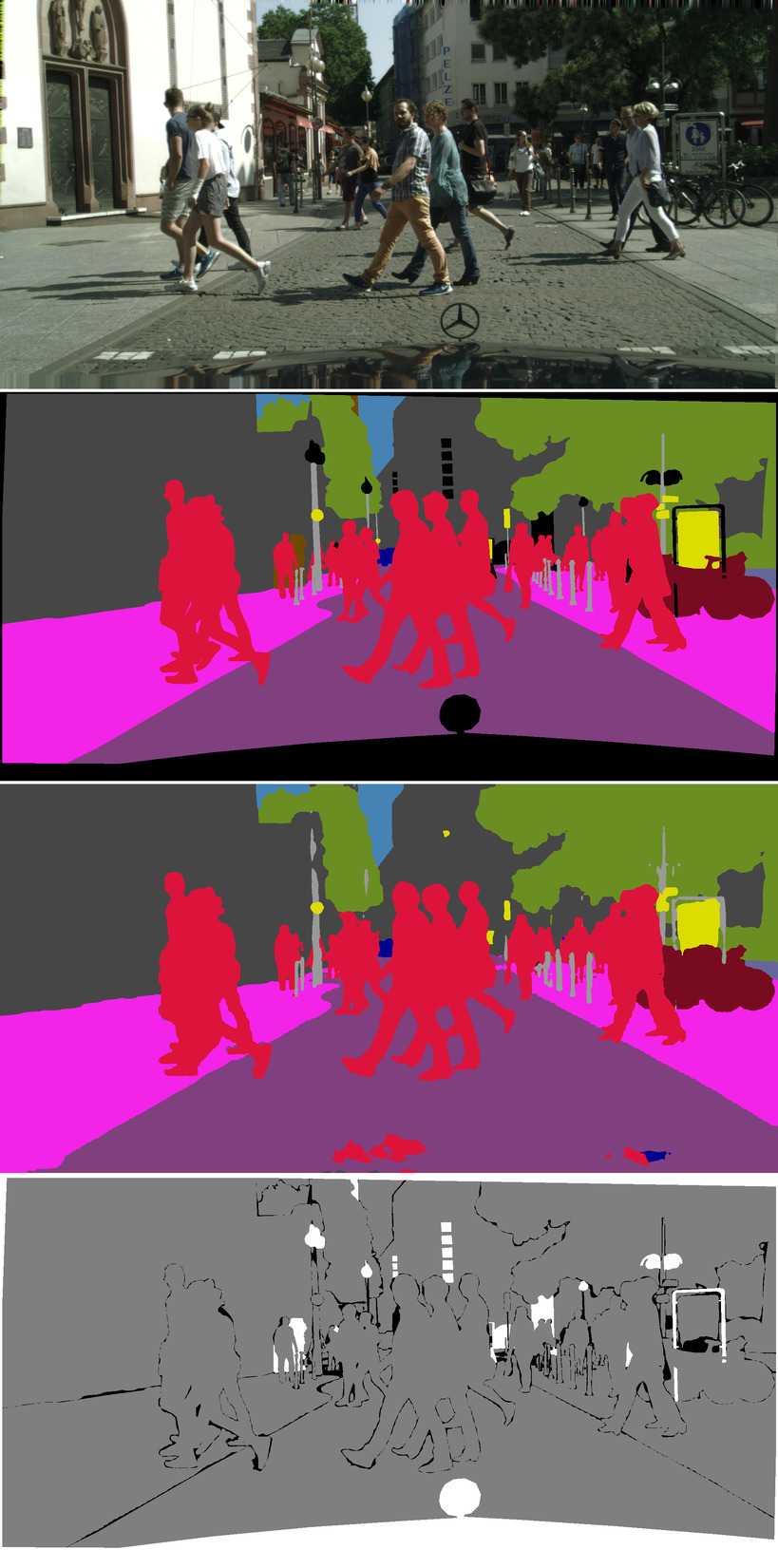}
\includegraphics[width=0.30\linewidth,trim=0 0 0 0]{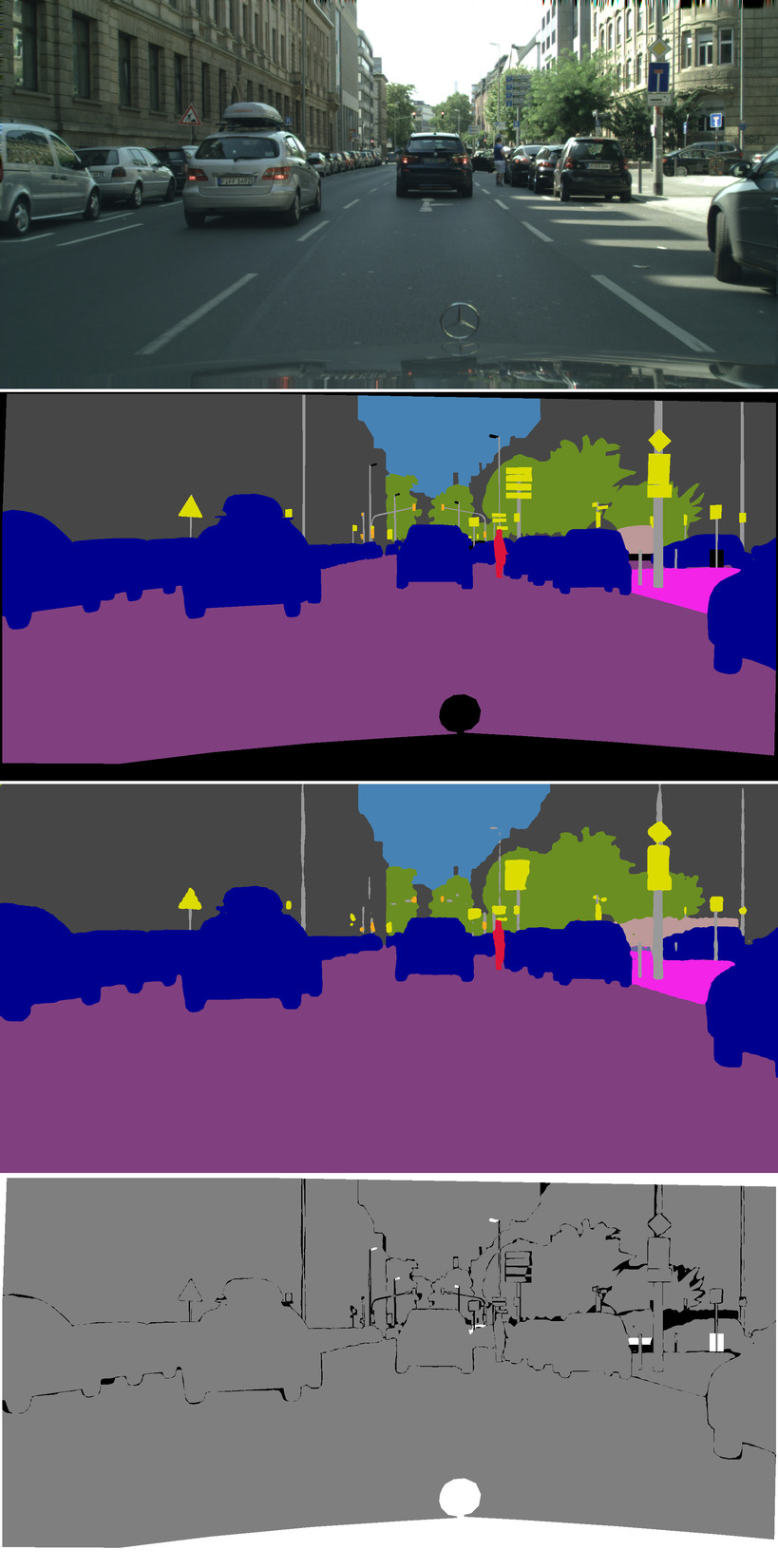}
\includegraphics[width=0.30\linewidth,trim=0 0 0 0]{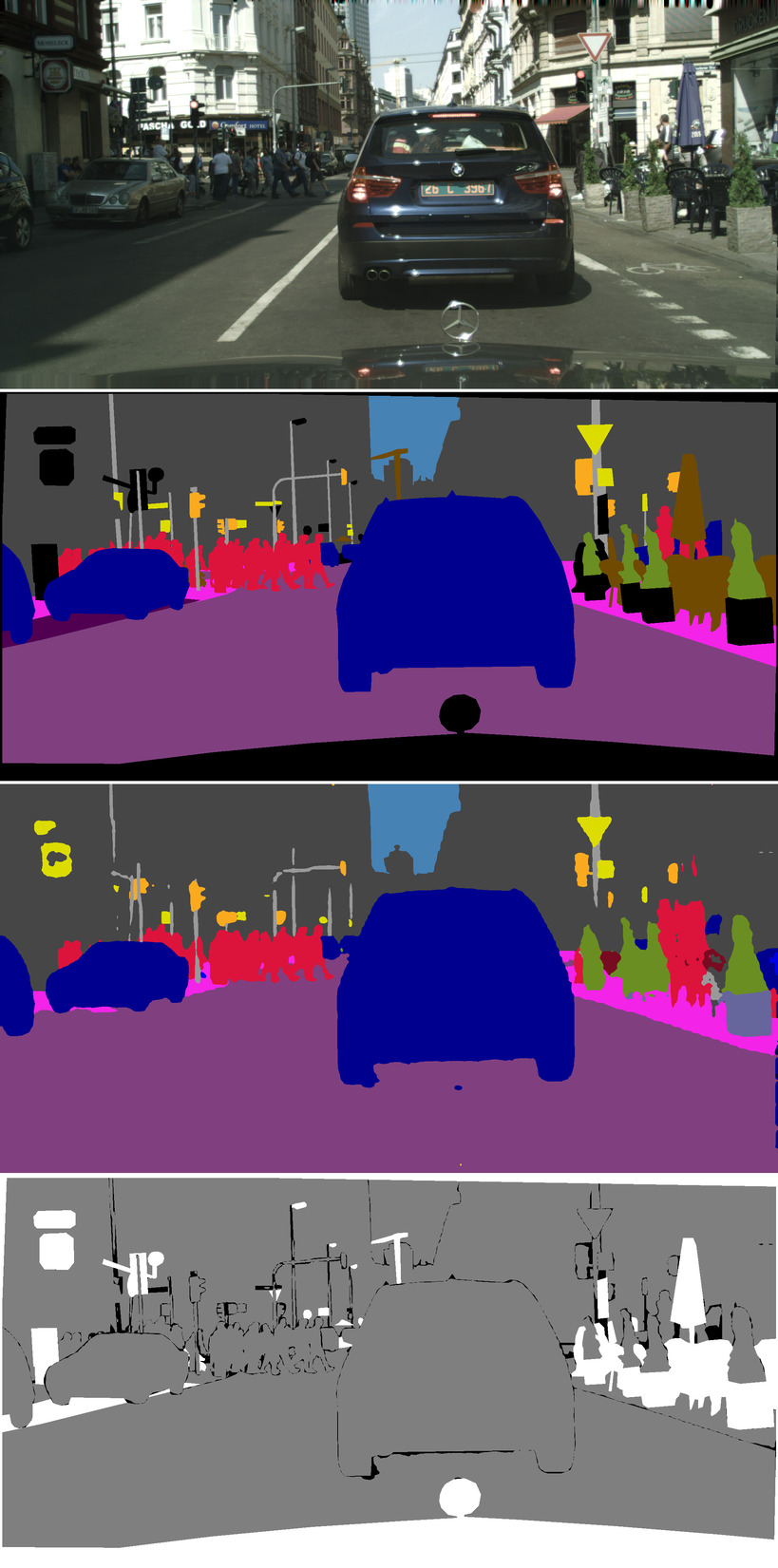}
\includegraphics[width=0.30\linewidth,trim=0 0 0 0]{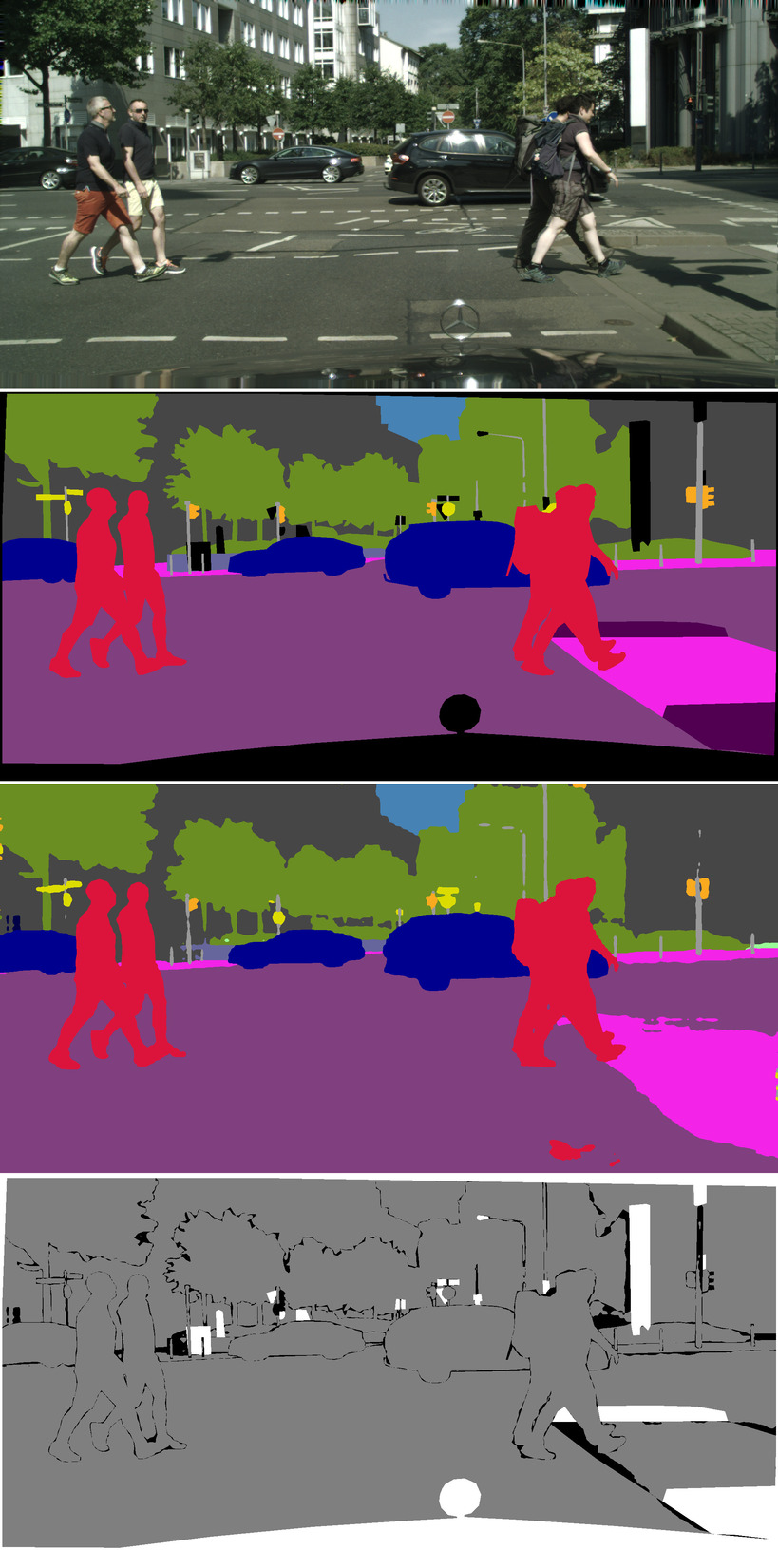}
\end{center}
\caption{
Qualitative results on the Cityscapes~\cite{Cityscapes.CVPR.2016.Cordts} val set.
The model was trained using the train set.
In each example, from top to bottom, there are in turn the original image, the ground-truth, the predicted label, and the difference map between the ground-truth and the predicted label.
}
\label{fig:cityscapes}
\end{figure*}

\begin{figure*}[t]
\vspace{-8.0mm}
\begin{center}
\includegraphics[width=0.15\linewidth,trim=0 0 0 0]{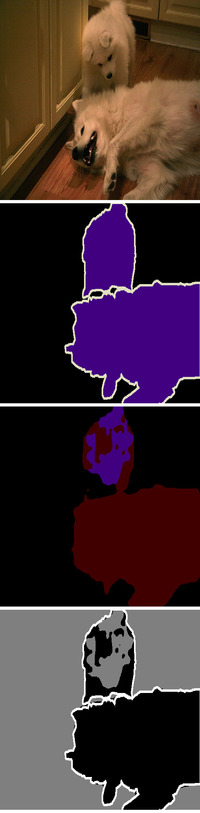}
\includegraphics[width=0.15\linewidth,trim=0 0 0 0]{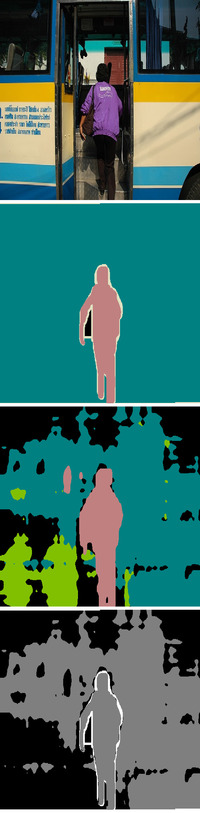}
\includegraphics[width=0.15\linewidth,trim=0 0 0 0]{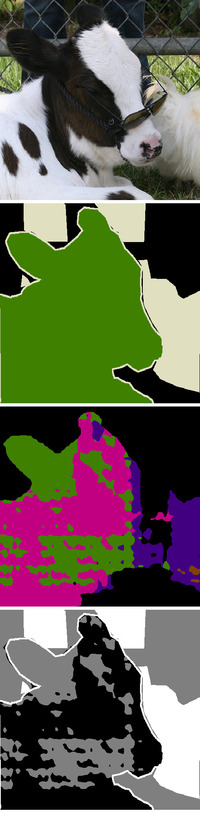}
\includegraphics[width=0.15\linewidth,trim=0 0 0 0]{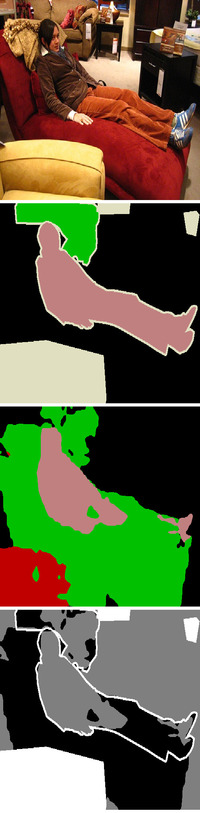}
\includegraphics[width=0.15\linewidth,trim=0 0 0 0]{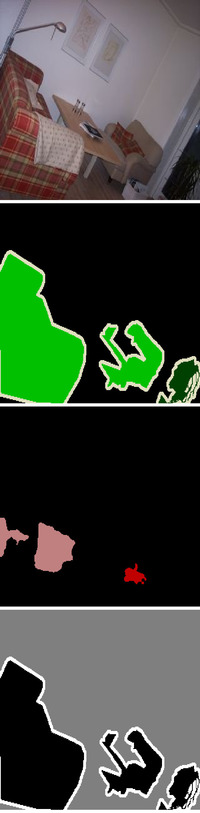}
\includegraphics[width=0.15\linewidth,trim=0 0 0 0]{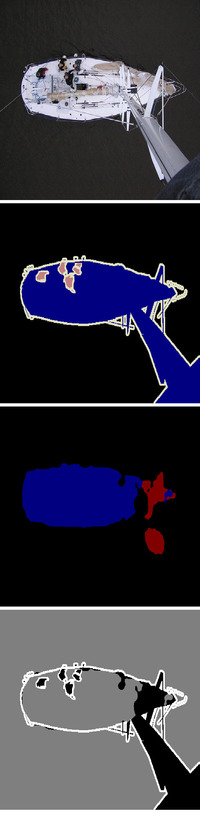}
\end{center}
\vspace{-4.0mm}
\caption{
Failure cases on the PASCAL VOC 2012~\cite{PascalVoc.IJCV.2014.Everingham} val set.
The model was trained using the train set augmented using SBD~\cite{SBD.ICCV.2011.Hariharan}.
In each example, from top to bottom, there are in turn the original image, the ground-truth, the predicted label, and the difference map between the ground-truth and the predicted label.
}
\label{fig:voc fail}
\vspace{-2.0mm}
\end{figure*}

\begin{figure*}[t]
\begin{center}
\includegraphics[width=0.30\linewidth,trim=0 0 0 0]{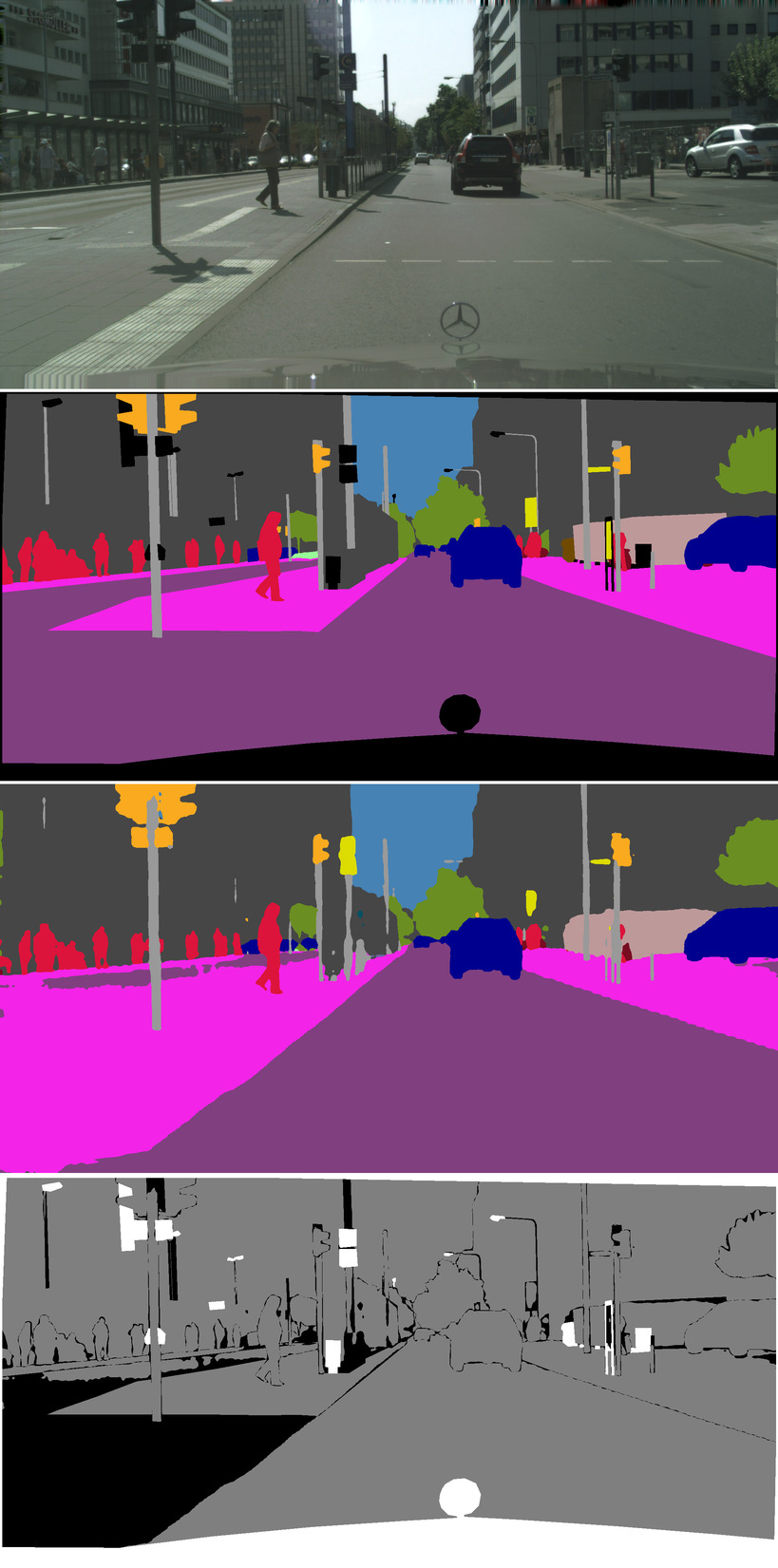}
\includegraphics[width=0.30\linewidth,trim=0 0 0 0]{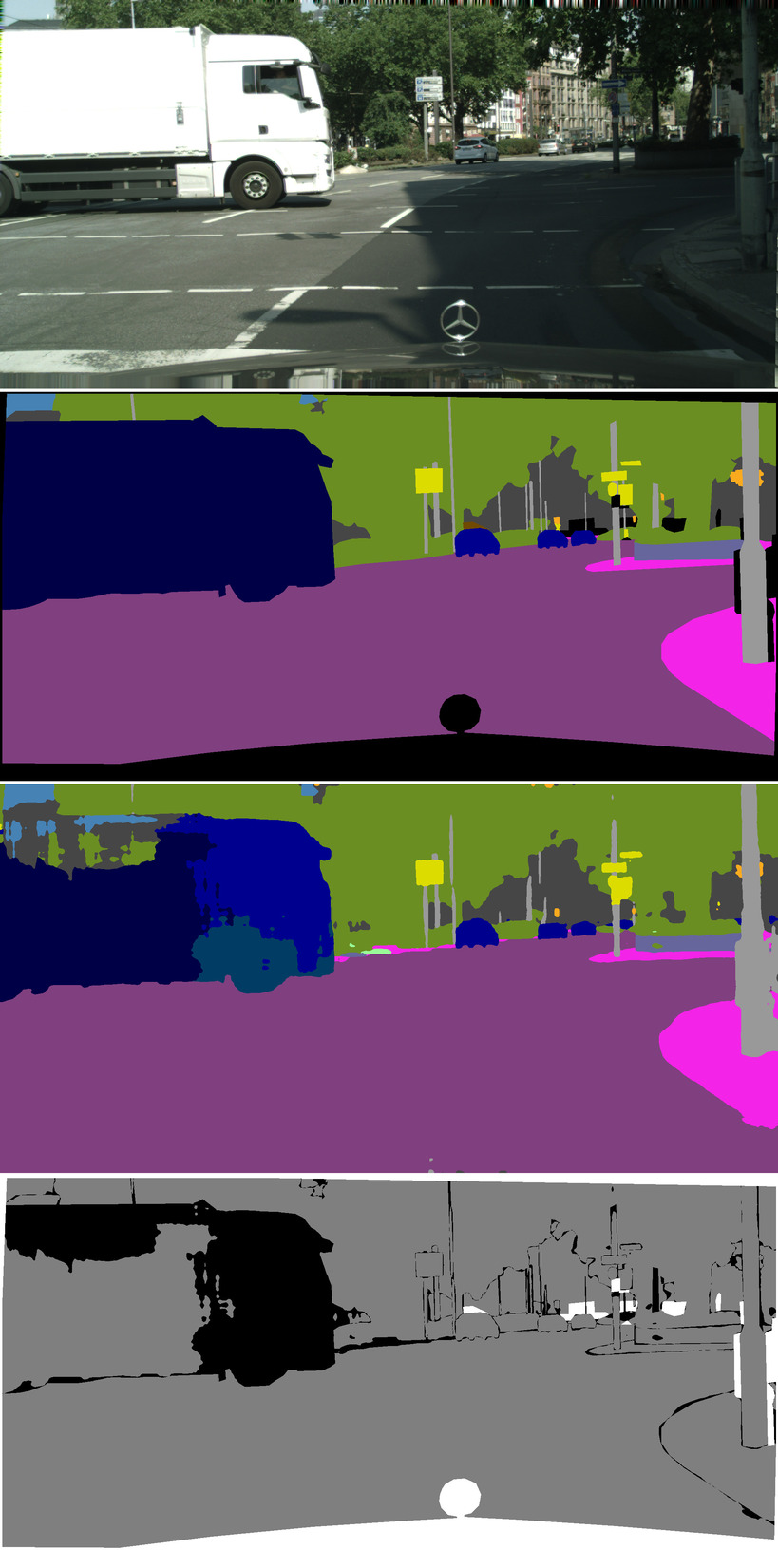}
\includegraphics[width=0.30\linewidth,trim=0 0 0 0]{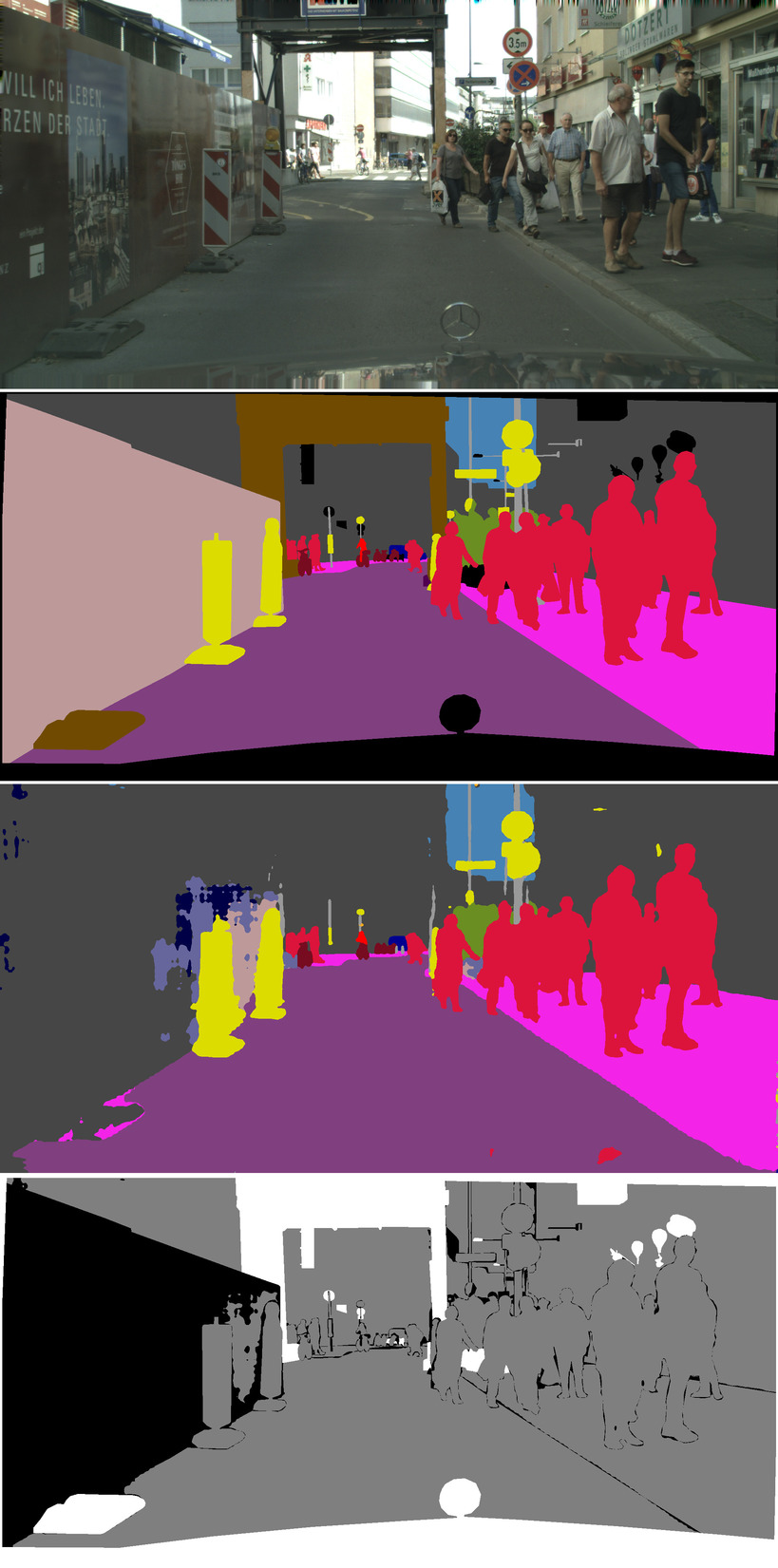}
\end{center}
\vspace{-4.0mm}
\caption{
Failure cases on the Cityscapes~\cite{Cityscapes.CVPR.2016.Cordts} val set.
The model was trained using the train set.
In each example, from top to bottom, there are in turn the original image, the ground-truth, the predicted label, and the difference map between the ground-truth and the predicted label.
}
\label{fig:cityscapes fail}
\vspace{-4.0mm}
\end{figure*}

\begin{figure*}[h]
\begin{center}
\includegraphics[width=0.15\linewidth,trim=0 0 0 0]{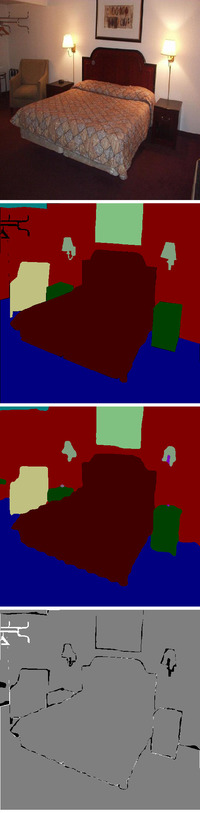}
\includegraphics[width=0.15\linewidth,trim=0 0 0 0]{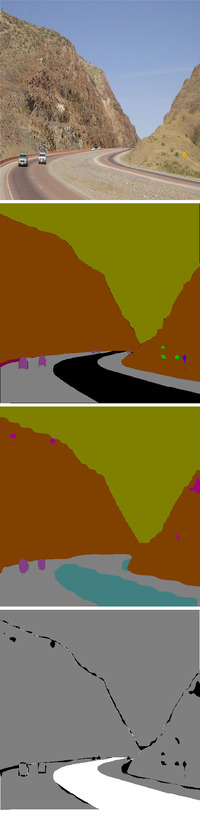}
\includegraphics[width=0.15\linewidth,trim=0 0 0 0]{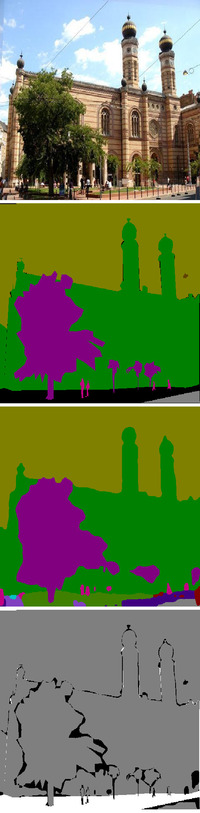}
\includegraphics[width=0.15\linewidth,trim=0 0 0 0]{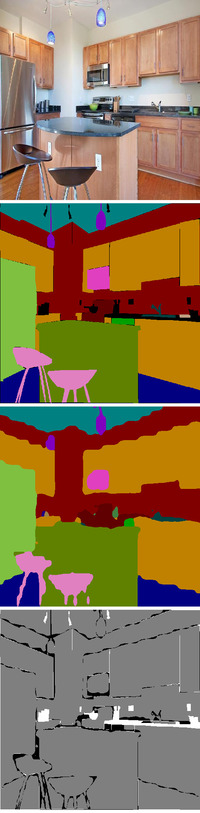}
\includegraphics[width=0.15\linewidth,trim=0 0 0 0]{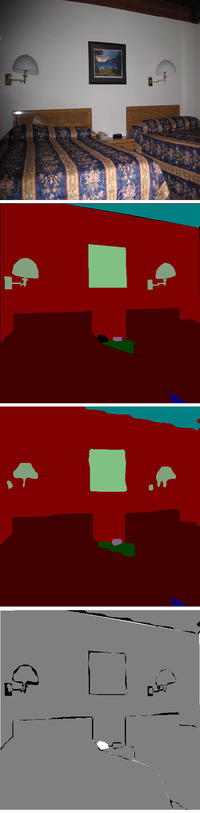}
\includegraphics[width=0.15\linewidth,trim=0 0 0 0]{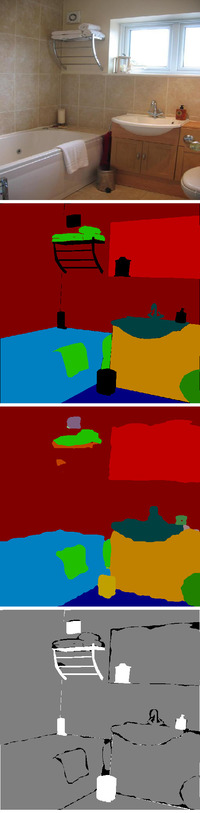}
\includegraphics[width=0.15\linewidth,trim=0 0 0 0]{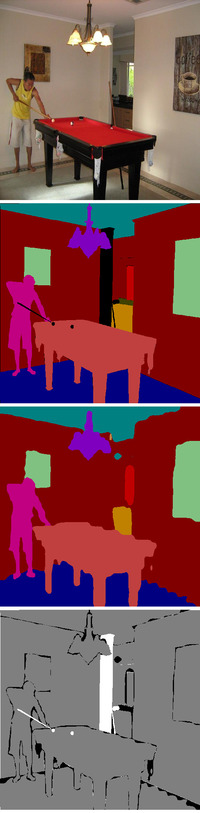}
\includegraphics[width=0.15\linewidth,trim=0 0 0 0]{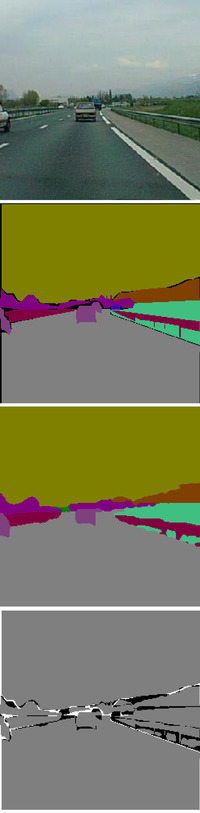}
\includegraphics[width=0.15\linewidth,trim=0 0 0 0]{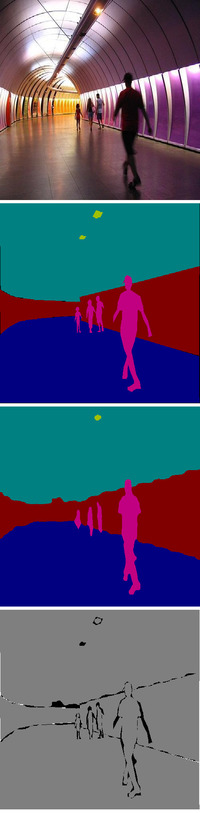}
\includegraphics[width=0.15\linewidth,trim=0 0 0 0]{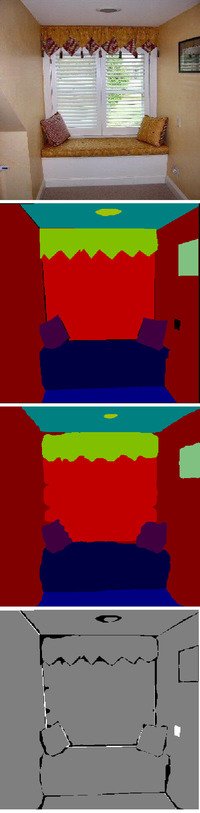}
\includegraphics[width=0.15\linewidth,trim=0 0 0 0]{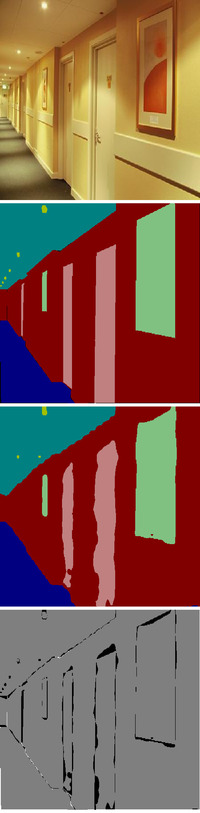}
\includegraphics[width=0.15\linewidth,trim=0 0 0 0]{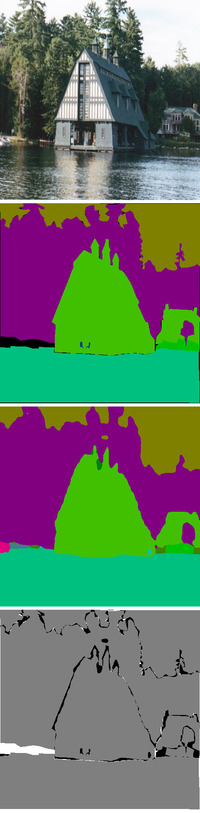}
\end{center}
\vspace{-5.0mm}
\caption{
Qualitative results on the ADE20K~\cite{ADE20K.2016.Zhou} val set.
The model was trained using the train set.
In each example, from top to bottom, there are in turn the original image, the ground-truth, the predicted label, and the difference map between the ground-truth and the predicted label.
}
\label{fig:ade20k}
\end{figure*}

\begin{figure*}[h]
\begin{center}
\includegraphics[width=0.15\linewidth,trim=0 0 0 0]{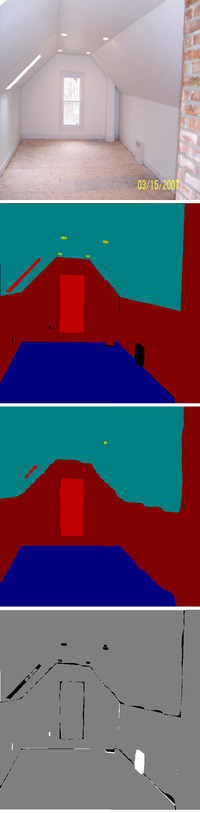}
\includegraphics[width=0.15\linewidth,trim=0 0 0 0]{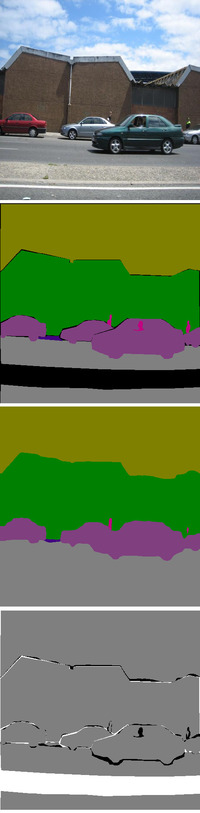}
\includegraphics[width=0.15\linewidth,trim=0 0 0 0]{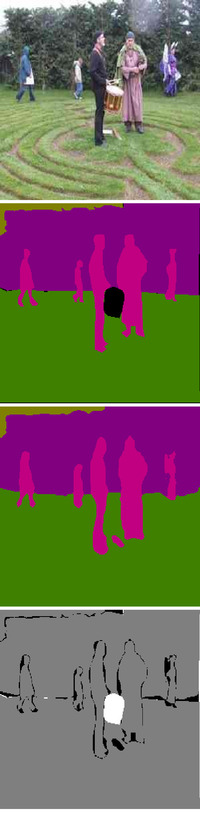}
\includegraphics[width=0.15\linewidth,trim=0 0 0 0]{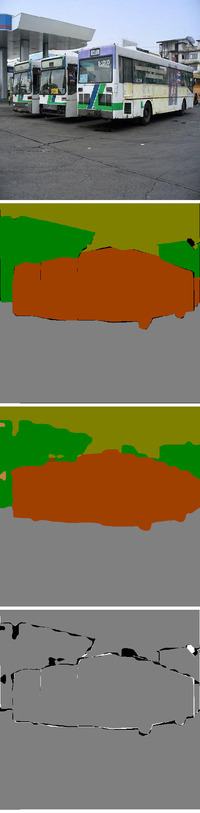}
\includegraphics[width=0.15\linewidth,trim=0 0 0 0]{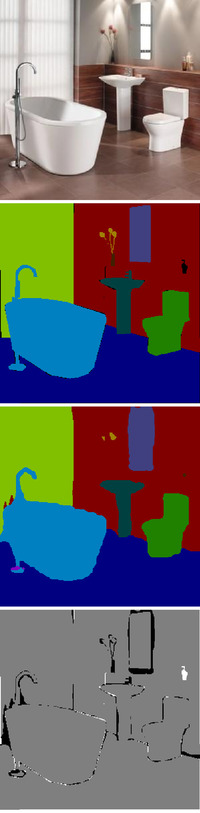}
\includegraphics[width=0.15\linewidth,trim=0 0 0 0]{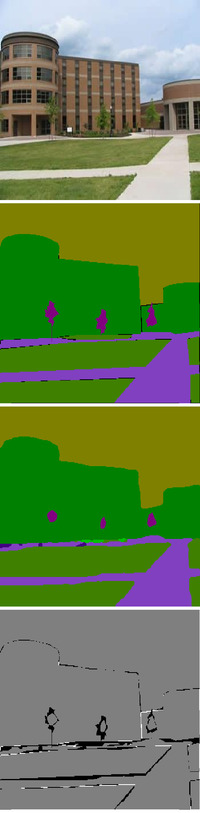}
\includegraphics[width=0.15\linewidth,trim=0 0 0 0]{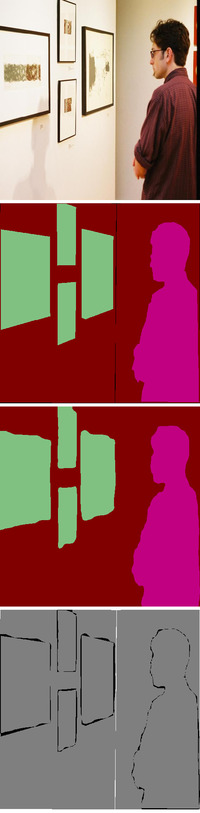}
\includegraphics[width=0.15\linewidth,trim=0 0 0 0]{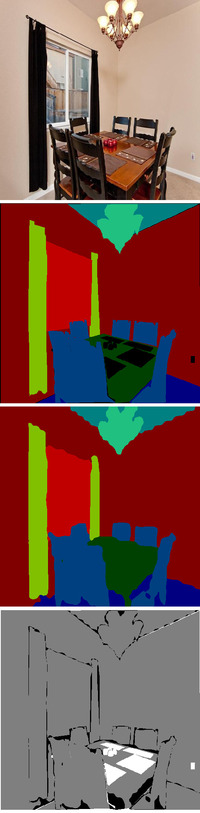}
\includegraphics[width=0.15\linewidth,trim=0 0 0 0]{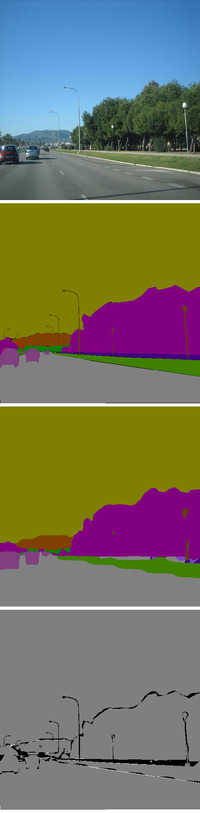}
\includegraphics[width=0.15\linewidth,trim=0 0 0 0]{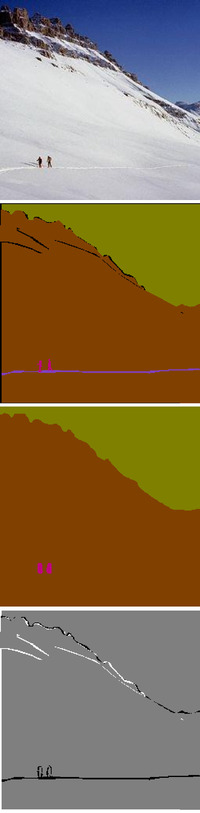}
\includegraphics[width=0.15\linewidth,trim=0 0 0 0]{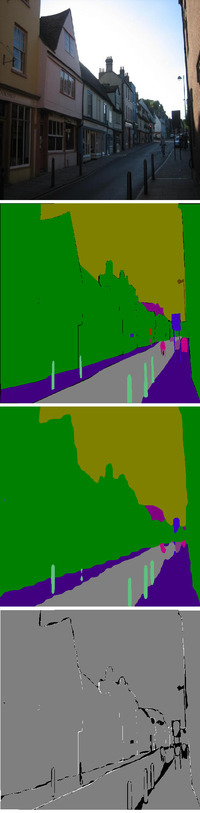}
\includegraphics[width=0.15\linewidth,trim=0 0 0 0]{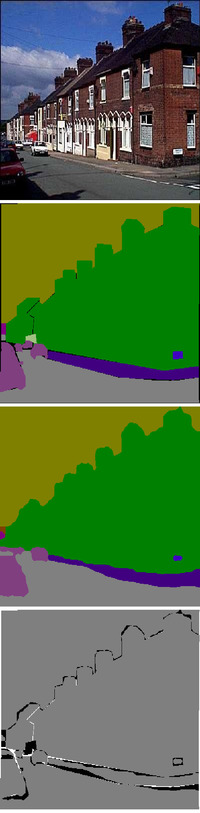}
\end{center}
\caption{
More qualitative results on the ADE20K~\cite{ADE20K.2016.Zhou} val set.
The model was trained using the train set.
In each example, from top to bottom, there are in turn the original image, the ground-truth, the predicted label, and the difference map between the ground-truth and the predicted label.
}
\label{fig:ade20k 2}
\end{figure*}

\begin{figure*}[h]
\begin{center}
\includegraphics[width=0.15\linewidth,trim=0 0 0 0]{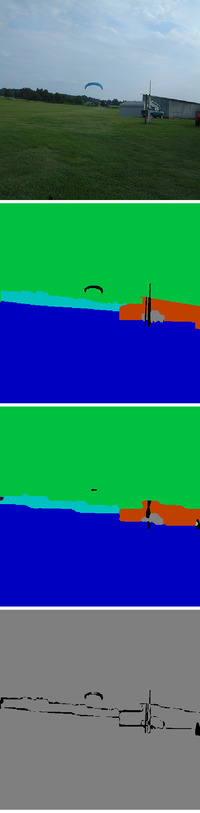}
\includegraphics[width=0.15\linewidth,trim=0 0 0 0]{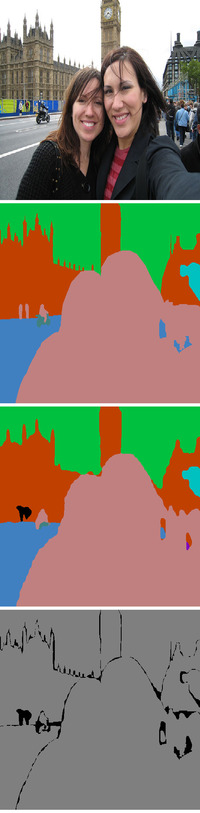}
\includegraphics[width=0.15\linewidth,trim=0 0 0 0]{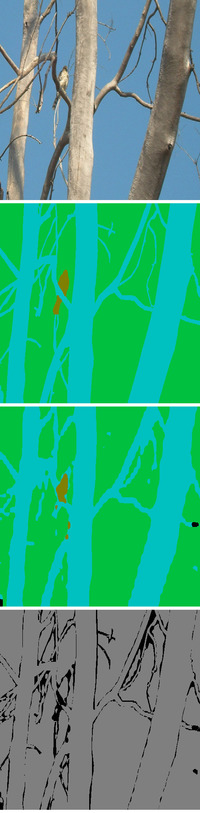}
\includegraphics[width=0.15\linewidth,trim=0 0 0 0]{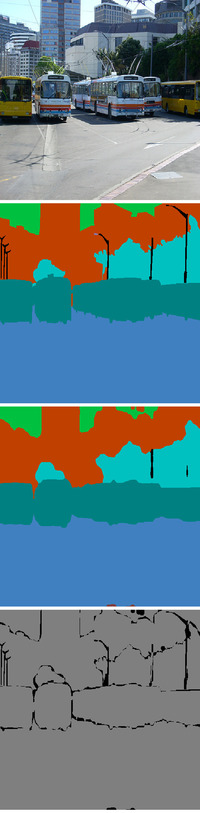}
\includegraphics[width=0.15\linewidth,trim=0 0 0 0]{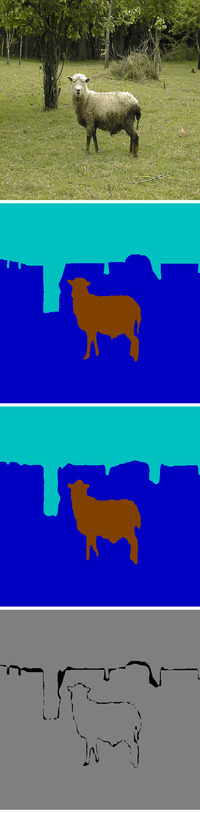}
\includegraphics[width=0.15\linewidth,trim=0 0 0 0]{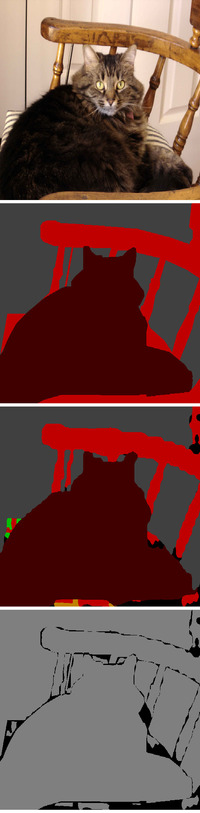}
\includegraphics[width=0.15\linewidth,trim=0 0 0 0]{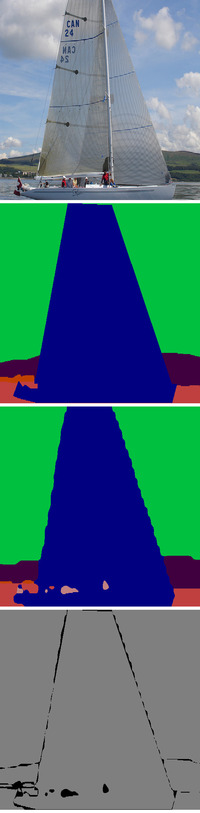}
\includegraphics[width=0.15\linewidth,trim=0 0 0 0]{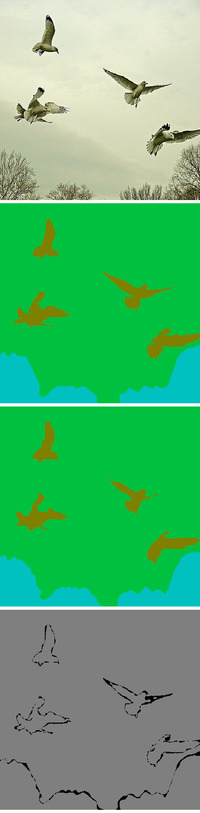}
\includegraphics[width=0.15\linewidth,trim=0 0 0 0]{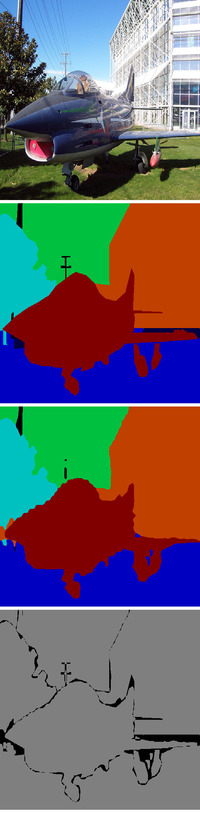}
\includegraphics[width=0.15\linewidth,trim=0 0 0 0]{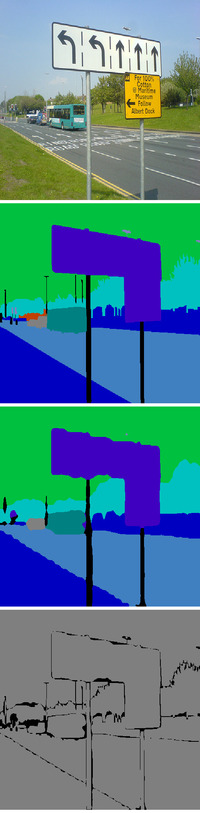}
\includegraphics[width=0.15\linewidth,trim=0 0 0 0]{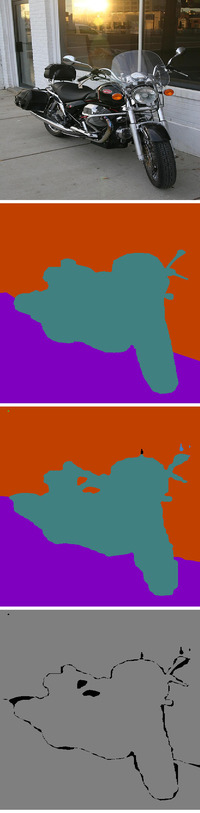}
\includegraphics[width=0.15\linewidth,trim=0 0 0 0]{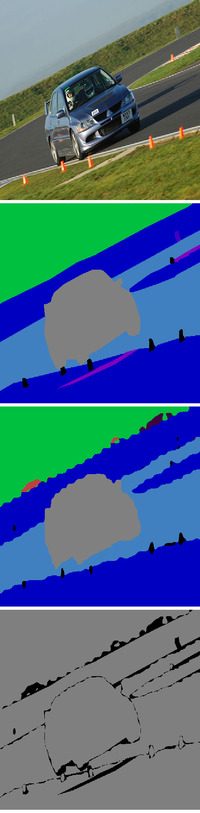}
\end{center}
\caption{
Qualitative results on the PASCAL Context~\cite{PascalContext.CVPR.2014.Mottaghi} val set.
The model was trained using the train set.
In each example, from top to bottom, there are in turn the original image, the ground-truth, the predicted label, and the difference map between the ground-truth and the predicted label.
}
\label{fig:pascal-context}
\end{figure*}

We show qualitative results of semantic image segmentation on PASCAL VOC~\cite{PascalVoc.IJCV.2014.Everingham}, Cityscapes~\cite{Cityscapes.CVPR.2016.Cordts}, ADE20K~\cite{ADE20K.2016.Zhou}, and PASCAL Context~\cite{PascalContext.CVPR.2014.Mottaghi},
respectively in Figs.~\ref{fig:voc},~\ref{fig:cityscapes},~\ref{fig:ade20k},~\ref{fig:ade20k 2}~and~\ref{fig:pascal-context},
and show some failure cases in Figs~\ref{fig:voc fail}~and~\ref{fig:cityscapes fail}.
In a difference map, grey and black respectively denotes correctly and wrongly labelled pixels,
while white denotes the officially ignored pixels during evaluation.
Note that we do not apply post-processing with CRFs,
which can smooth the output but is too slow in practice, especially for large images.

